\documentclass[preprint,5p,authoryear]{elsarticle}

\usepackage{amssymb}
\usepackage{natbib}
\usepackage{xspace}
\usepackage{appendix}
\usepackage{caption}  
\usepackage{subcaption}
\usepackage{multirow}
\usepackage{amsmath}
\usepackage{pgfplots}
\usepackage{graphicx}
\pgfplotsset{compat=1.18}
\usepackage[utf8]{inputenc}

\journal{ISPRS Journal of Photogrammetry and Remote Sensing}

\begin{document}

\begin{frontmatter}



\title{IMAFD: An Interpretable Multi-stage Approach to Flood Detection from time series Multispectral Data}

\author[label1]{Ziyang Zhang}
\author[label1]{ Plamen Angelov}
\author[label1]{ Dmitry Kangin}
\affiliation[label1]{organization={LIRA Centre, School of Computing and Communications},
            addressline={Lancaster University},
            city={Lancaster},
            postcode={LA1 4WA},
            country={UK}}
\author[label2]{Nicolas Long\'ep\'e}    
\affiliation[label2]{organization={$\Phi$-lab, European Space Agency},
            addressline={ESRIN},
            city={Frascati},
            postcode={00044},
            country={Italy}}

% \author{Ziyang Zhang ^a, Plamen Angelov ^a, Dmitry Kangin ^a, Nicolas Longepe, Pierre Philippe Mathieu and Giuseppe Borghi}

% \affiliation{organization={School of Computing and Communications, Lancaster University},%Department and Organization
%             city={Lancaster},
%             postcode={LA1 4WA}, 
%             country={UK}}

\begin{abstract}
In this paper, we address two critical challenges in the domain of flood detection: the computational expense of large-scale time series change detection and the lack of interpretable decision-making processes on explainable AI (XAI). To overcome these challenges, we proposed an interpretable multi-stage approach to flood detection, IMAFD has been proposed. It provides an automatic, efficient and interpretable solution suitable for large-scale remote sensing tasks and offers insight into the decision-making process. The proposed IMAFD approach combines the analysis of the dynamic time series image sequences to identify images with possible flooding with the static, within-image semantic segmentation. It combines anomaly detection (at both image and pixel level) with semantic segmentation. The flood detection problem is addressed through four stages: (1) at a sequence level: identifying the suspected images (2) at a multi-image level: detecting change within suspected images (3) at an image level: semantic segmentation of images into Land, Water or Cloud class (4) decision making. Our contributions are two folder. First, we efficiently reduced the number of frames to be processed for dense change detection by providing a multi-stage holistic approach to flood detection. Second, the proposed semantic change detection method (stage 3) provides human users with an interpretable decision-making process, while most of the explainable AI (XAI) methods provide post hoc explanations. The evaluation of the proposed IMAFD framework was performed on three datasets, WorldFloods, RavAEn and MediaEval. For all the above datasets, the proposed framework demonstrates a competitive performance compared to other methods offering also interpretability and insight.
\end{abstract}


% \begin{highlights}
% \end{highlights}


\end{frontmatter}


\section{Introduction}

Floods are one of the most frequent and widespread natural disasters on a global scale \citep{zhao2023siam, iqbal2021computer}. The magnitude of their impact has escalated significantly in recent decades. From 2000 to 2019, the population affected by floods increased from 58 million to 86 million. This statistic exhibits a tenfold increase compared to earlier estimations. Forecasts indicate that a significantly increased proportion of the population will be exposed to floods \citep{jongman2012global, tellman2021satellite}. 

To mitigate and address the hazards posed by floods, remote sensing technologies have been used. Unlike traditional field-based instruments, they have the advantage of providing near real-time and large-area monitoring. Therefore, they have become a powerful tool for Earth Observation (EO) tasks, especially for large-scale flood extent monitoring. In the area of remote sensing, both synthetic aperture radar (SAR) and multispectral optical images are widely used. SAR has problems with noise and information loss \citep{zhang2021unsupervised}. Optical images take advantage of the fact that different materials reflect and absorb differently in different wavebands, and found extensive applications in EO tasks, such as flood mapping \citep{mateo2021towards} and change detection \citep{ruuvzivcka2022ravaen}.

To date, various change detection and semantic segmentation algorithms have been developed and introduced to the flood detection problem. The traditional change detection and semantic segmentation methods normally rely on threshold-based \citep{long2014flood, schlaffer2015flood} or classic-machine-learning-based \citep{landuyt2020flood, pan2020comparative} methods. However, such methods usually have the disadvantage of not providing high accuracy. Recent studies utilizing deep learning based methods \citep{khelifi2020deep, mateo2021towards}, despite providing promising results, such methods usually have two disadvantages, Firstly, such change detection methods or semantic segmentation methods usually require making dense predictions to every frame of the time series images, which is computationally expense. Besides, such models usually contain millions or even billions of trainable parameters which are hard to interpret. This makes it difficult for human users to understand the decision-making process of such algorithms.  

To address such issues, this paper presents a Multi-stage Approach to Flood Detection (IMAFD). The inspiration for this is that in a time series of images, the occurrence of a flood is a minority event. In such cases, identifying floods from such time series of images can be considered as an anomaly detection problem. After novel images are detected, binary change detection and semantic change detection are performed only on novel images to determine the location and type of pixels where the anomaly occurred. Finally, the information obtained from the above steps is used in the decision-making step to find out whether flooding has occurred or not.  Overall, IMAFD provides an efficient, and human-understandable flood detection framework by dividing flood detection in four steps. The contribution of our work can be summarised as follows.

\begin{figure*}[h!]
\centerline{\includegraphics[width=0.9\textwidth]{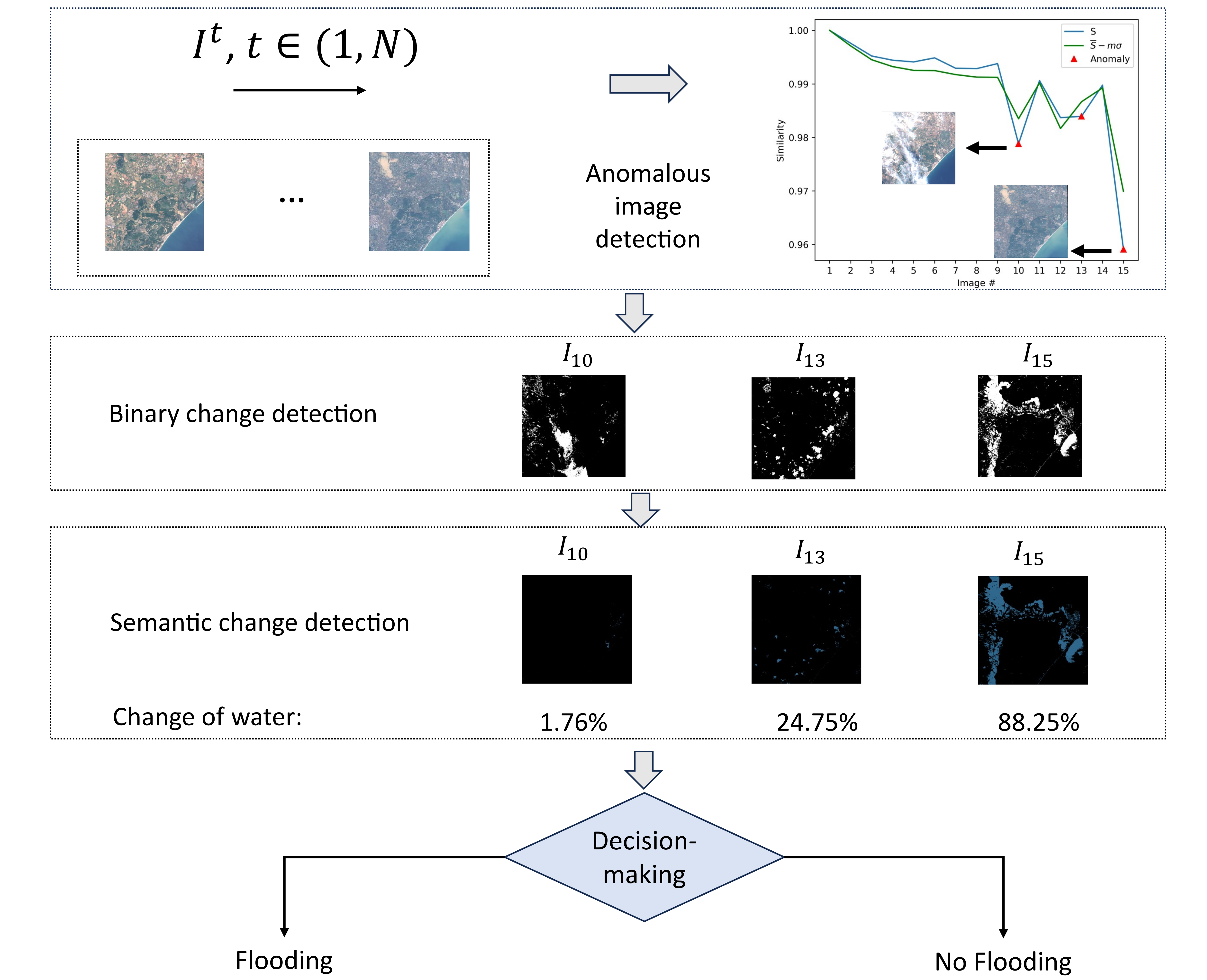}}
\caption{IMAFD architecture. The proposed IMAFD architecture consists of the following four stages, (1) anomalous image detection: identifying suspected images from time series multispectral optical images (2) binary change detection: detecting changes within suspected images (3) semantic change detection: semantic segmentation of changes into Land, Water or Cloud class (4) decision making.}
\label{fig1}
\end{figure*}

1)	The IMAFD framework is proposed by treating flood detection as a multi-stage problem. It efficiently reduces the number of images that need to be processed for dense prediction in the later stage by performing anomaly detection in the first stage, which is computationally friendly.

2) The IDSS+ is introduced as a critically important component of the IMAFD framework. Compared to IDSS \citep{zhang2022interpretable}, IDSS+ improves the performance and provides an option for better interpretability by utilizing real pixels as prototypes at the expense of a small reduction in performance. In addition, several techniques are introduced, that help human users to better understand the decision-making process of the algorithm. These include UMAP plots and confidence maps, in addition to the linguistic form of prototypes-based explanation which IDSS already had.

3)	The IMAFD architecture including IDSS+ is validated on the change detection dataset RaVAEn \citep{ruuvzivcka2022ravaen}, anomaly detection dataset MediaEval and the semantic segmentation dataset WorldFloods \citep{mateo2021towards}, respectively. The proposed methods provide promising performance in terms of numerical results and interpretability.

\section{Related work}

\subsection{Semantic Segmentation on flood detection}

Traditional semantic segmentation methods mainly rely on threshold-based \citep{al2010image} and clustering-based \citep{zheng2018image} methods. Specifically for flood detection tasks, water-index-based methods are particularly popular, such as Normalized Difference Water Index (NDWI) \citep{gao1996ndwi}, Modified Normalized Difference Water Index (MNDWI) \citep{xu2006modification}, and Normalized Difference Vegetation Index (NDVI), etc. However, the performance of such methods is highly dependent on expert knowledge and constrained by various limitations. For example, NDWI might mix Water and Land, while the optical threshold for MNDWI could not remove the shadow noise in some regions. Most importantly, such methods usually require careful selection of the optimal threshold value, which has to be adjusted according to different regions and times. All these drawbacks greatly hinder the application of such methods in global-scale flood mapping \citep{zhou2017open}.

In recent years, with the growth of visual data and the development of hardware devices, deep learning algorithms have been widely used in the area of computer vision, including semantic segmentation. In particular, the introduction of the Fully Convolutional Network \citep{long2015fully} is considered as a milestone in the field of semantic segmentation, since it is the first framework that implements end-to-end semantic segmentation by using deep learning techniques. Following this, several well-known semantic segmentation networks have been proposed, such as U-Net \citep{ronneberger2015u}, SegNet \citep{badrinarayanan2017segnet}, and DeepLabV3+ \citep{chen2018encoder}.  More recently, with the success of the vision transformers thanks to their ability to provide state-of-the-art performance, which is facilitated by large-scale datasets and the model’s global content modeling capabilities\citep{dosovitskiy2020image}. Several transformer-based semantic segmentation neural networks have emerged, such as SETR \citep{zheng2021rethinking}, SegFormer \citep{xie2021segformer}, Segmenter \citep{strudel2021segmenter}, etc. Such deep learning-based models usually employ an encoder-decoder architecture, where the encoder typically uses a convolutional neural network (CNN) or transformer to extract hierarchical features by progressively downsampling the input, while the decoder upsamples and fuses such features to get the final pixel-wise segmentation map \citep{zheng2021rethinking}. Overall, these deep learning-based semantic segmentation models are broadly used in various domains due to their ability to provide encouraging performance.

For example, numerous deep learning-based algorithms were developed and applied to EO tasks, including flood extent mapping. For instance, \citep{mateo2021towards} applied U-Net to Sentinel-2 images for flood mapping, while \citep{katiyar2021near} employed SegNet and U-Net for flood mapping with SAR images. \citep{wu2022flood} proposed a novel multi-scale DeepLab model based on MobileNetV2 backbone and DeepLabv3+ for flood detection with SAR images. \citep{zhou2022water} introduced a CNN-transformer-based model applied to SAR images, demonstrating improved performance of fine water region detection.
Despite their advantages, deep learning algorithms have been criticized for their “black box” nature \citep{rudin2019stop, angelov2020towards, angelov2023towards}.

\subsection{Change detection}

In the remote sensing field, most change detection methods are based on bi-temporal images, typically identifying changes that occur in the same region during a time interval. Most traditional bi-temporal change detection methods normally use image differences or ratios based on the raw features \citep{mall2022change}. For example, Change Vector Analysis (CVA) \citep{malila1980change} employed the difference between bi-temporal images for calculating change vector while \citep{gong2011neighborhood} introduces a novel neighbourhood-based ratio operator to generate the change map. Recent work has mostly used deep learning-based methods \citep{ bai2023deep, chen2021remote} thanks to their powerful feature extraction capabilities and promising performance. However, bi-temporal change detection methods are usually difficult to deal with images taken from different seasons and tend to introduce seasonal errors in the results \citep{liu2015sequential, hao2023bi}. 

To address this problem, some researchers have proposed time series change detection methods, which could effectively reduce the seasonal error and obtain a large amount of land use information \citep{ zhu2017change}. However, such methods usually need to perform classification or segmentation for each image, which is time-consuming and inefficient\citep{yan2019time}.

\subsection{Explainable AI}
As deep learning techniques are widely used in various industries, more and more people are concerned about the limitations of these algorithms, with a particular focus on transparency and explainability \citep{ gevaert2022explainable}. The necessity for interpretable algorithms is not only due to the need to meet the legal and regulatory requirements of the governments and organizations \citep{regulation2016regulation}, but also to help the users understand and audit the decision-making process of the algorithms.

Despite the rapid development of deep learning, especially deep neural network techniques, they have been criticised for the reason of their “black box” nature. This fact has motivated researchers to start exploring interpretable algorithms, and many representative works have been presented. Early works focused on post hoc explanations. For example, GradCam \citep{ selvaraju2017grad} provides an interpretation of a concept by generating a coarse localization map in the final convolutional layer to highlight important regions associated with the concept. LIME \citep{ribeiro2016should} provides an explanation by constructing a simple linear model trained on a set of generated perturbed instances to find the highest weighted superpixels. These interpretable methods have been widely used in Earth observation. For example, \citep{guo2022prob} proposed a variant of GradCam, Prob-Cam, and successfully applied it to the remote sensing image classification task. Similarly, \citep{hung2020remote} applied the LIME method to the task of remote sensing scene classification. However, such class of explanations may not faithfully reflect the intrinsic decision-making. More recent work addresses the question of post hoc explanations \citep{chan2022comparative}, however, to entirely address these problems, one can consider interpretable-by-design methods.

Several interpretable-by-design models have been proposed. For example, Bayesian Neural Networks (BNNs) provide explanations based on uncertainty predictions that could help human users to understand how unsure the results are \citep{clare2022explainable, mitros2019validity}. B-cos networks are introduced to increase the interpretability of deep neural networks (DNNs) by promoting weight-input alignment during taring \citep{bohle2022b, bohle2024b}.  

In addition to the above, the prototype-based models have attracted attention for their natural proximity to human reasoning \citep{biehl2016prototype}. Related works include xDNN \citep{angelov2020towards}, ProtoPNet \citep{chen2019looks}, DNC \citep{ wang2022visual}, IDEAL \citep{ angelov2023towards} and IDSS \citep{ zhang2022interpretable}. All of these models are trained either by adding a prototype layer to the neural network and training the model in an end-to-end manner, or by performing a clustering algorithm on a trained/pre-trained feature space of the neural network to find the most representative prototypes.

Among these approaches, the work most similar to our proposed framework is xDNN \citep{angelov2020towards}, IDEAL \citep{ angelov2023towards} and IDSS \citep{ zhang2022interpretable}. However, the first two were used for the classification task of RGB images while the proposed framework is used for the semantic segmentation task with multispectral optical images. Compared to IDSS, the proposed framework further refines the approach by providing better performance and several techniques to help human users understand the algorithm’s decision-making process.

\begin{figure*}[htbp]
\centerline{\includegraphics[width=0.9\linewidth]{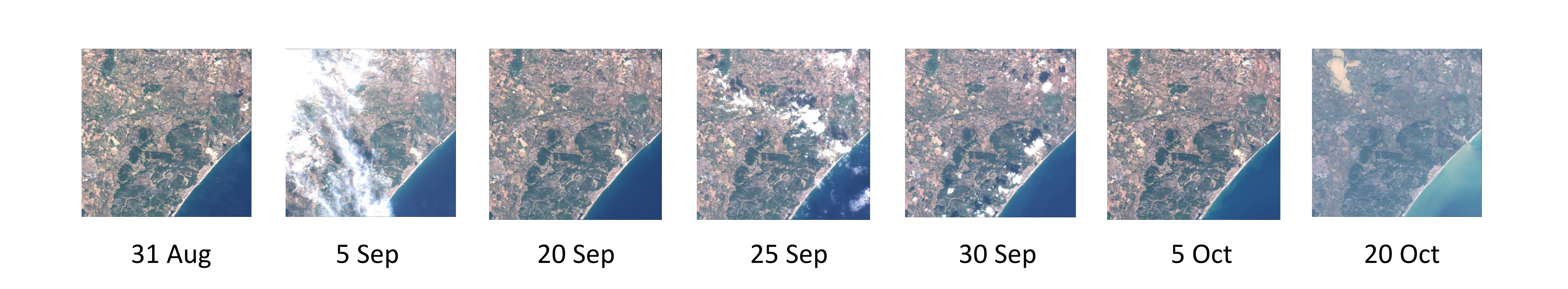}}
\caption{RGB visualization of Sentinel-2 time series multispectral images (All of the above images were taken in 2018).}
\label{time_series}
\end{figure*}

\section{Methodology}

\subsection{Problem statement}

In time series image analysis, flood events are often considered to be a type of low-frequency anomalous event, and thus time series flood detection can be categorised as a time series change detection task. Currently, most time series change detection algorithms need to generate pixel-level change maps for each image frame, which is not only time-consuming but also inefficient. In addition, due to the widespread use of deep learning methods, the decision-making process of such methods is often hard to explain and incomprehensible to human users.

To address these challenges, the IMAFD framework is proposed and introduced in this paper. IMAFD divides flood detection into four steps. First, IMAFD identifies anomalous images from a set of time series images. Recursive Similarity Estimation (RSE) based on the Recursive Density Estimation (RDE) \citep{angelov2012autonomous}  was used. It is specifically adapted to the flooding task and is detailed in the Appendix. Next, if the image is identified as an anomaly, a similarity-based binary change detection method is used to locate the anomalous region within that image. Furthermore, to classify such regions, the IDSS+ method is employed to distinguish between Land, Water, and Clouds. Finally, a decision-making process combines the above information to determine where the flooding occurred and the specific flooded areas within that image.

\begin{figure*}[h!]
\centerline{\includegraphics[width=0.9\linewidth]{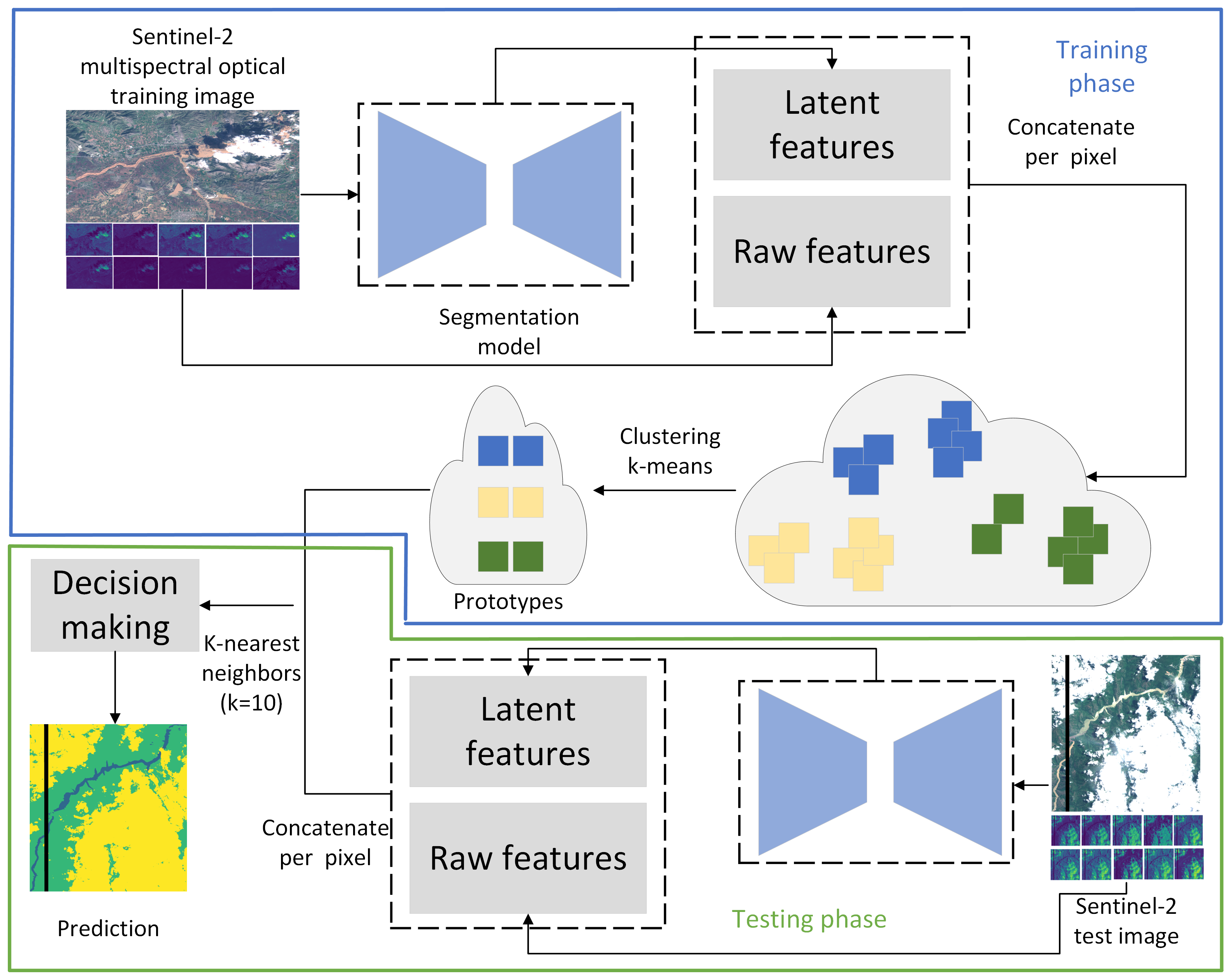}}
\caption{IDSS+ architecture.}
\label{IDSS+}
\end{figure*}

The overall architecture of the IMAFD framework is shown in Figure \ref{fig1}. The details of the proposed architecture and its function are introduced as follows.

\subsection{IMAFD framework}

\subsubsection{Anomalous images detection}

Figure \ref{time_series} shows a series of Sentinel-2 time series multispectral images that are visualized by the RGB channels, which tend to show anomalies such as cloud cover, natural disasters, urban development, and agricultural harvesting. Among these anomalies, this paper mainly focuses on flooding. In this context, a flood usually occurs when land is covered by floodwater, such as when cities or farmland are flooded, or when existing rivers or lakes begin to expand. Another situation that needs to be considered is that cloud cover is a common phenomenon in such time series multispectral images. In contrast, flooding is a low-probability event.

To detect anomalous images including clouds and floods, we use the statistics for outlier estimation by detecting drops in the overall image similarity degree, $S$, according to the following criterion:

\begin{equation}
\begin{aligned}
    IF \quad S^k < \Bar{S}^{k}-m\sigma^{k}\\
    THEN \quad (possible \quad novelty)\\
    ELSE \quad(normal \quad image)
\end{aligned}
\end{equation}

For $k = 1, ..., N$ images of the data stream, m is a positive number, $S \in (0;1]$ is the overall image similarity degree and $\Bar{S} \in (0;1]$ is the mean similarity degree.

\subsubsection{Binary Change detection}

\begin{figure*}[!h]
    \centering
    \captionsetup[subfloat]{labelformat=empty}
    
    \subfloat[]{\includegraphics[width=0.15\textwidth]{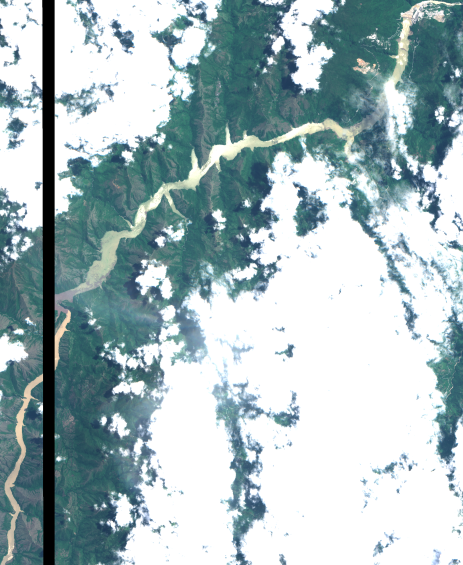}}\hfil
    \subfloat[]{\includegraphics[width=0.15\textwidth]{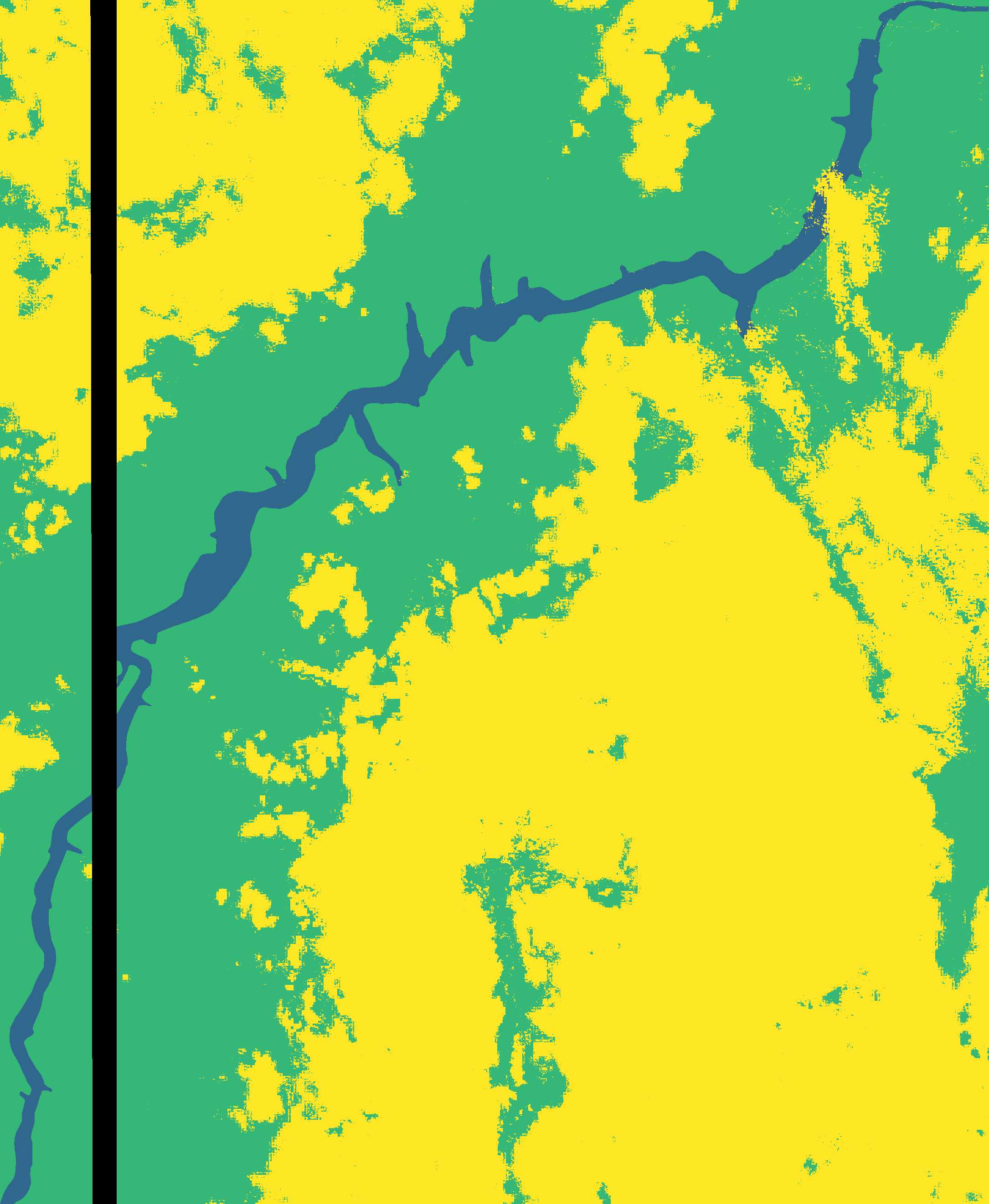}}\hfil
    \subfloat[]{\includegraphics[width=0.15\textwidth]{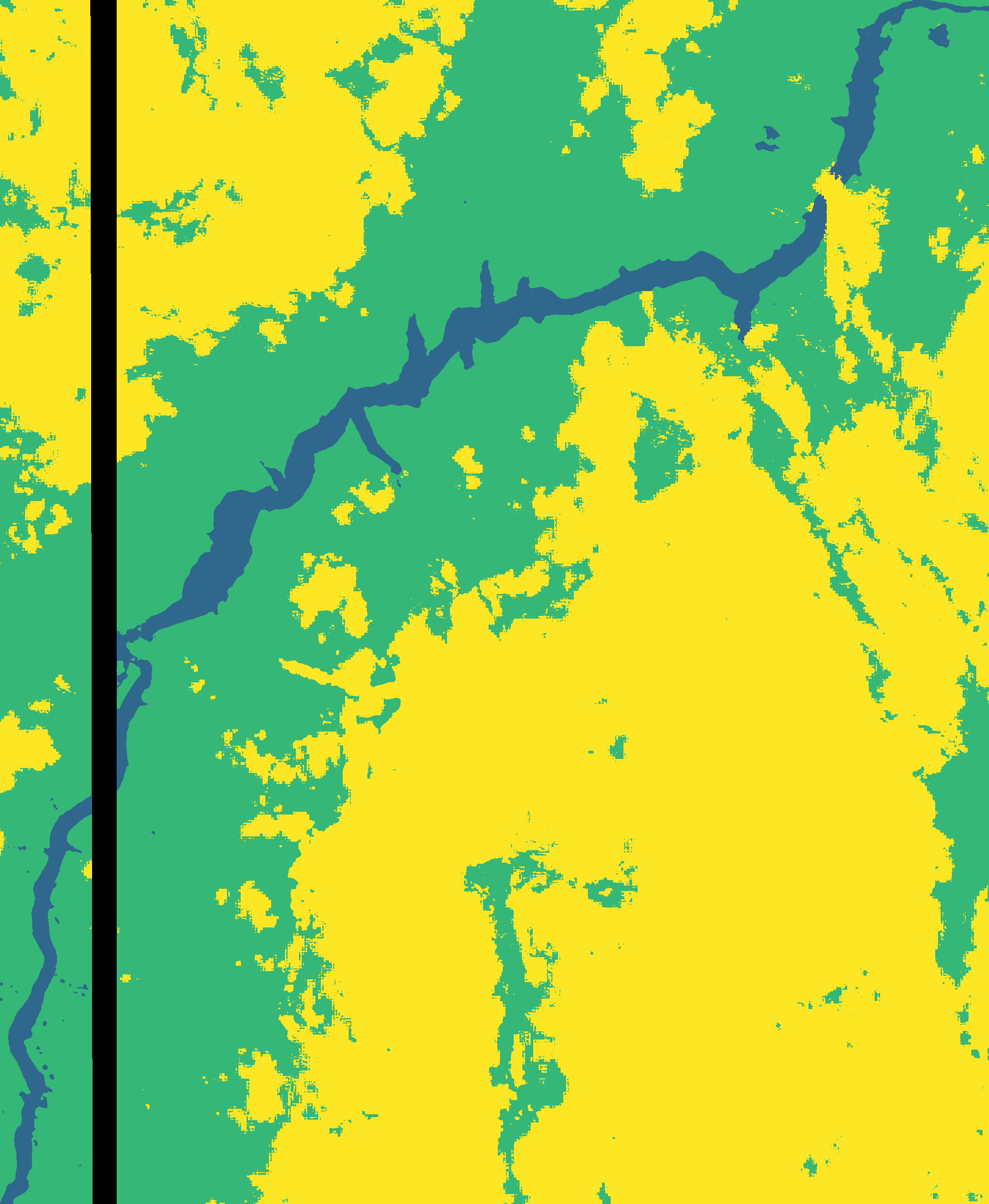}}\hfil
    \subfloat[]{\includegraphics[width=0.15\textwidth]{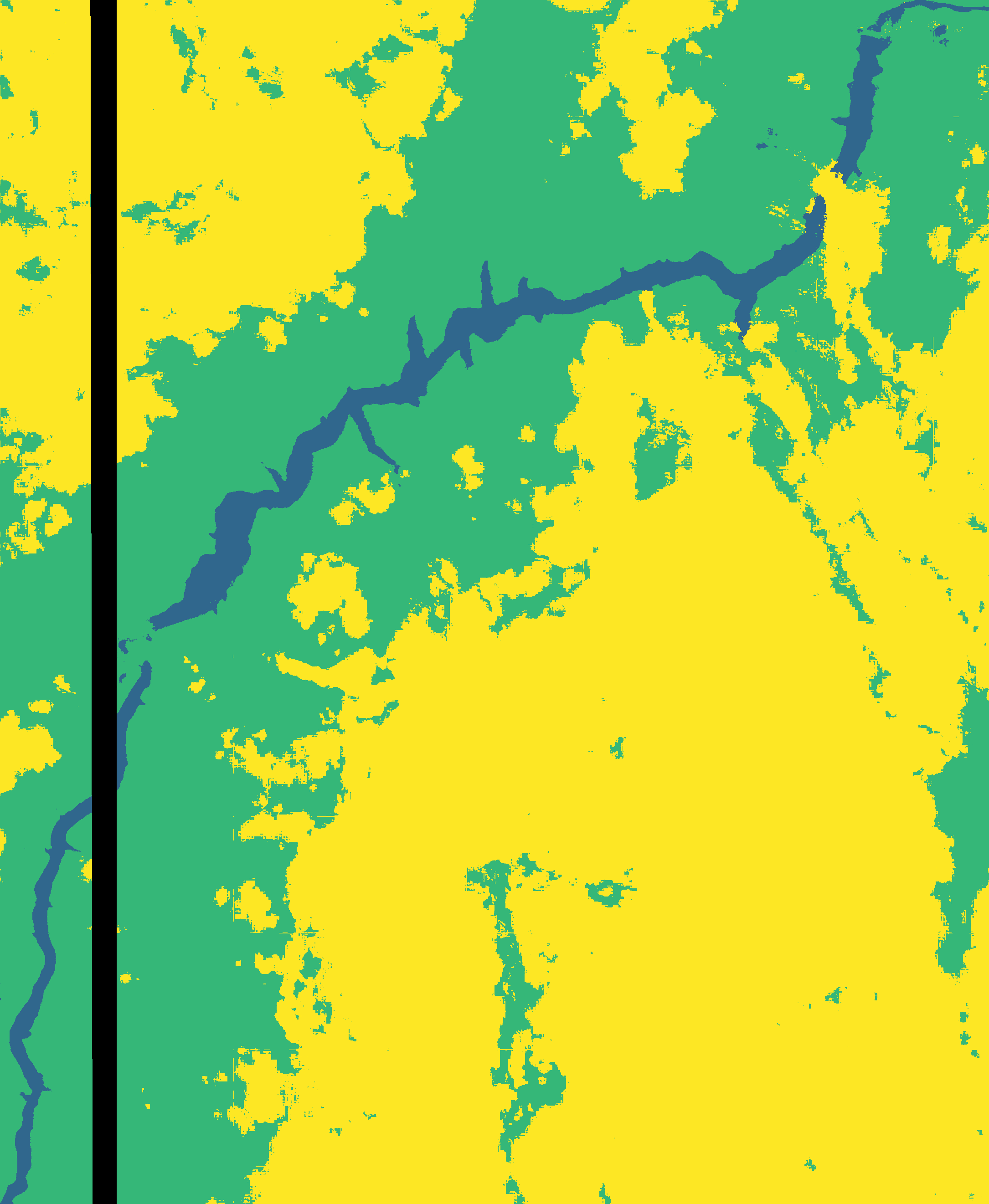}}\hfil
    \subfloat[]{\includegraphics[width=0.15\textwidth]{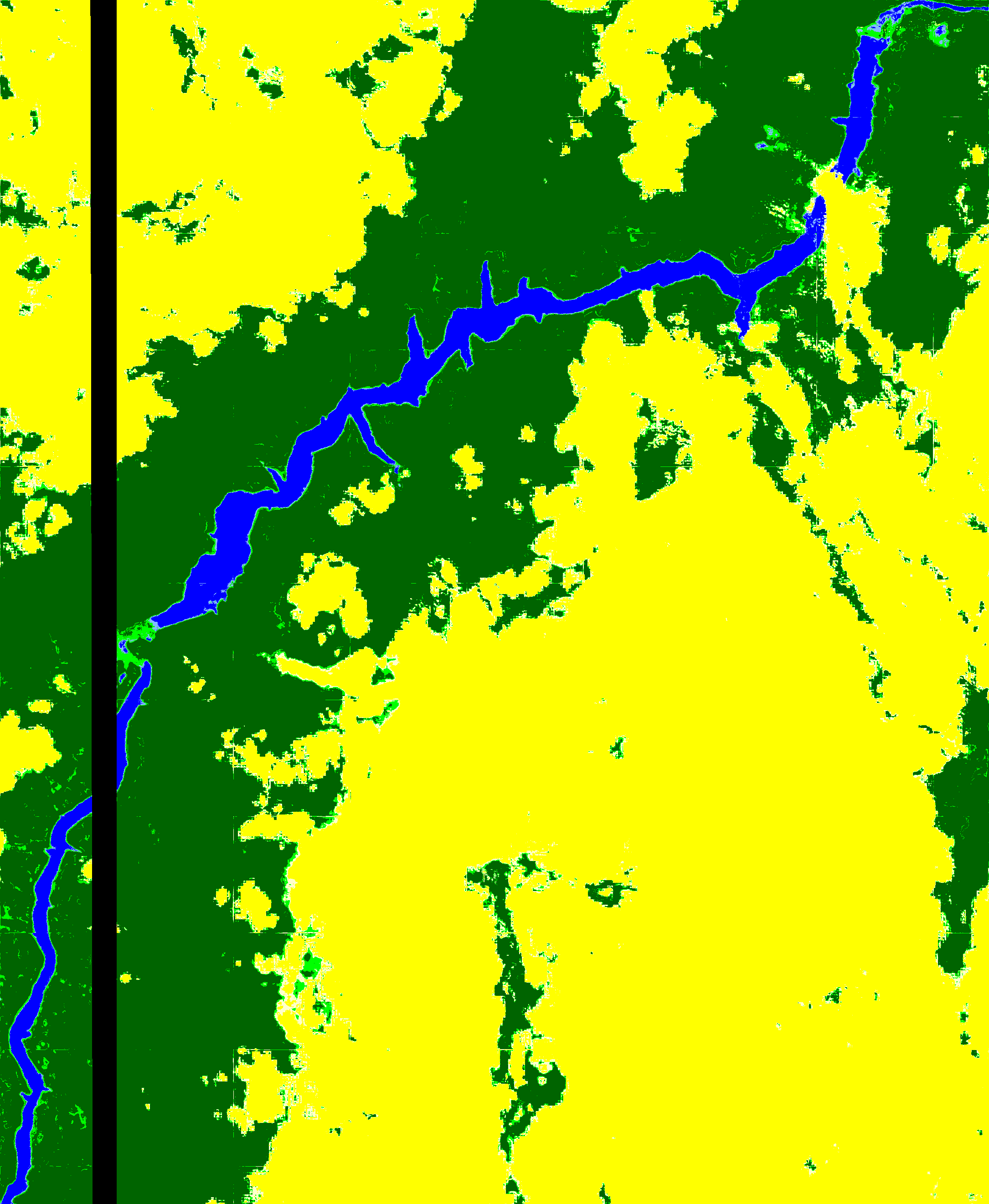}}\hfil
    
    \medskip
    \subfloat[]{\includegraphics[width=0.15\textwidth]{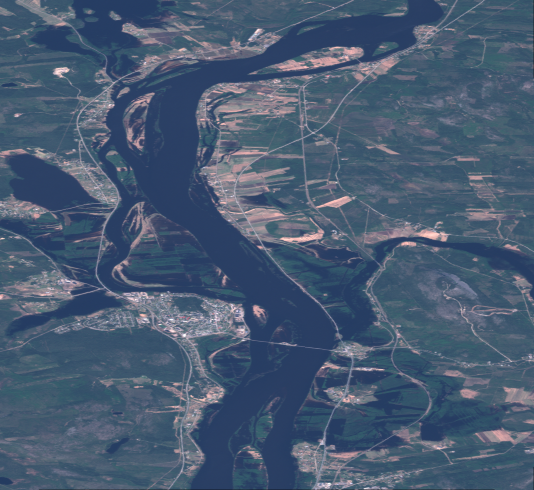}}\hfil
    \subfloat[]{\includegraphics[width=0.15\textwidth]{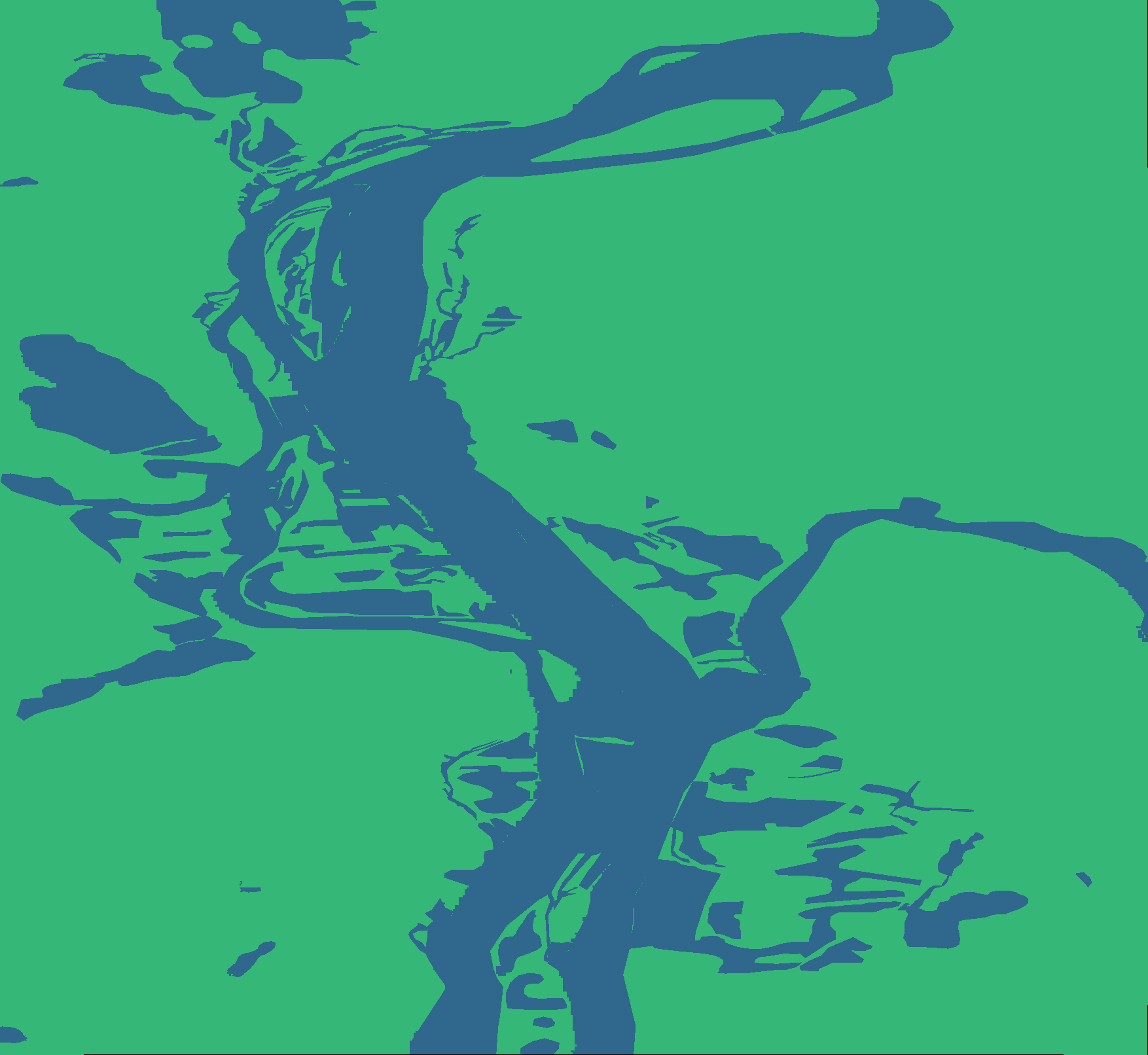}}\hfil
    \subfloat[]{\includegraphics[width=0.15\textwidth]{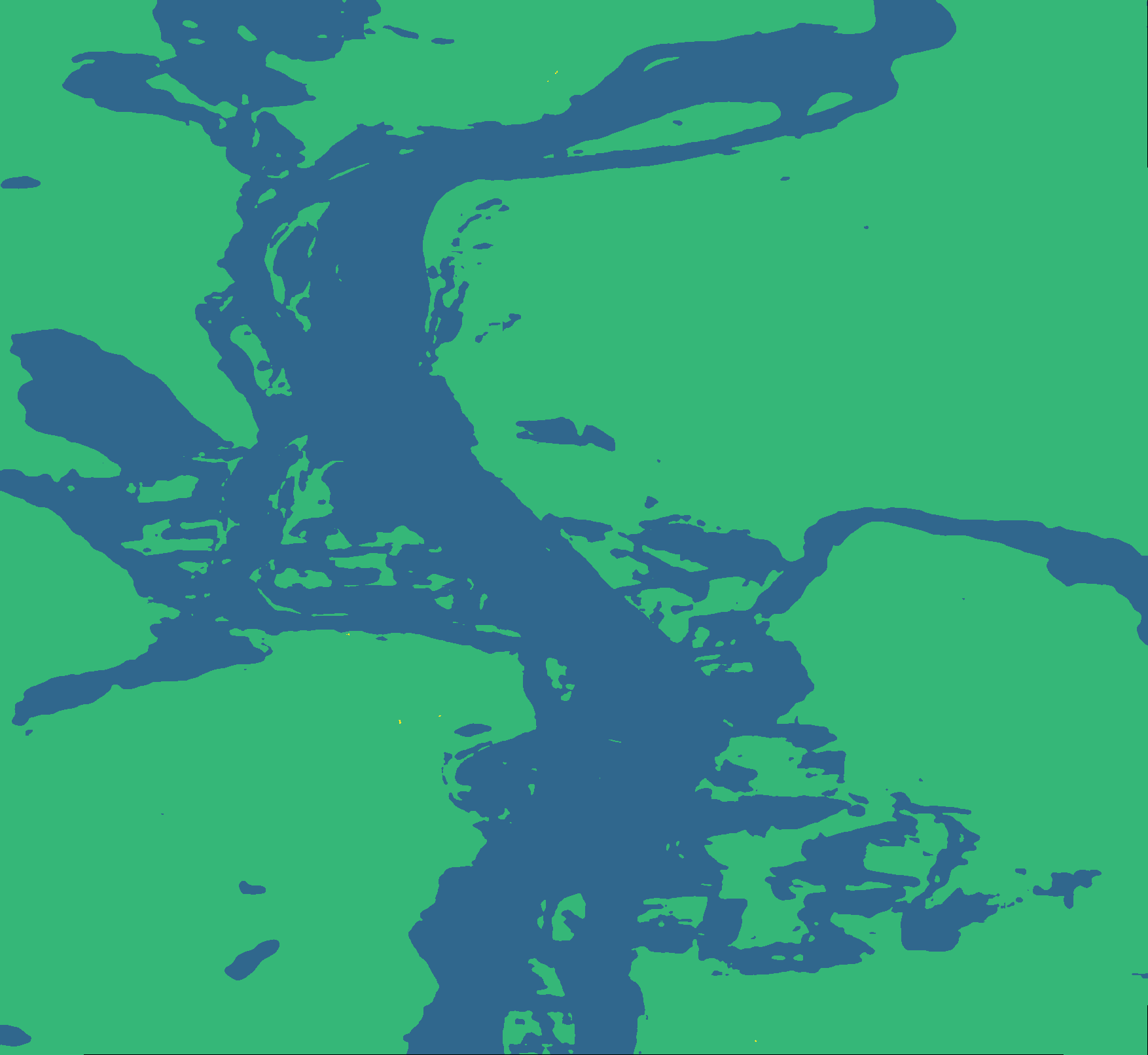}}\hfil
    \subfloat[]{\includegraphics[width=0.15\textwidth]{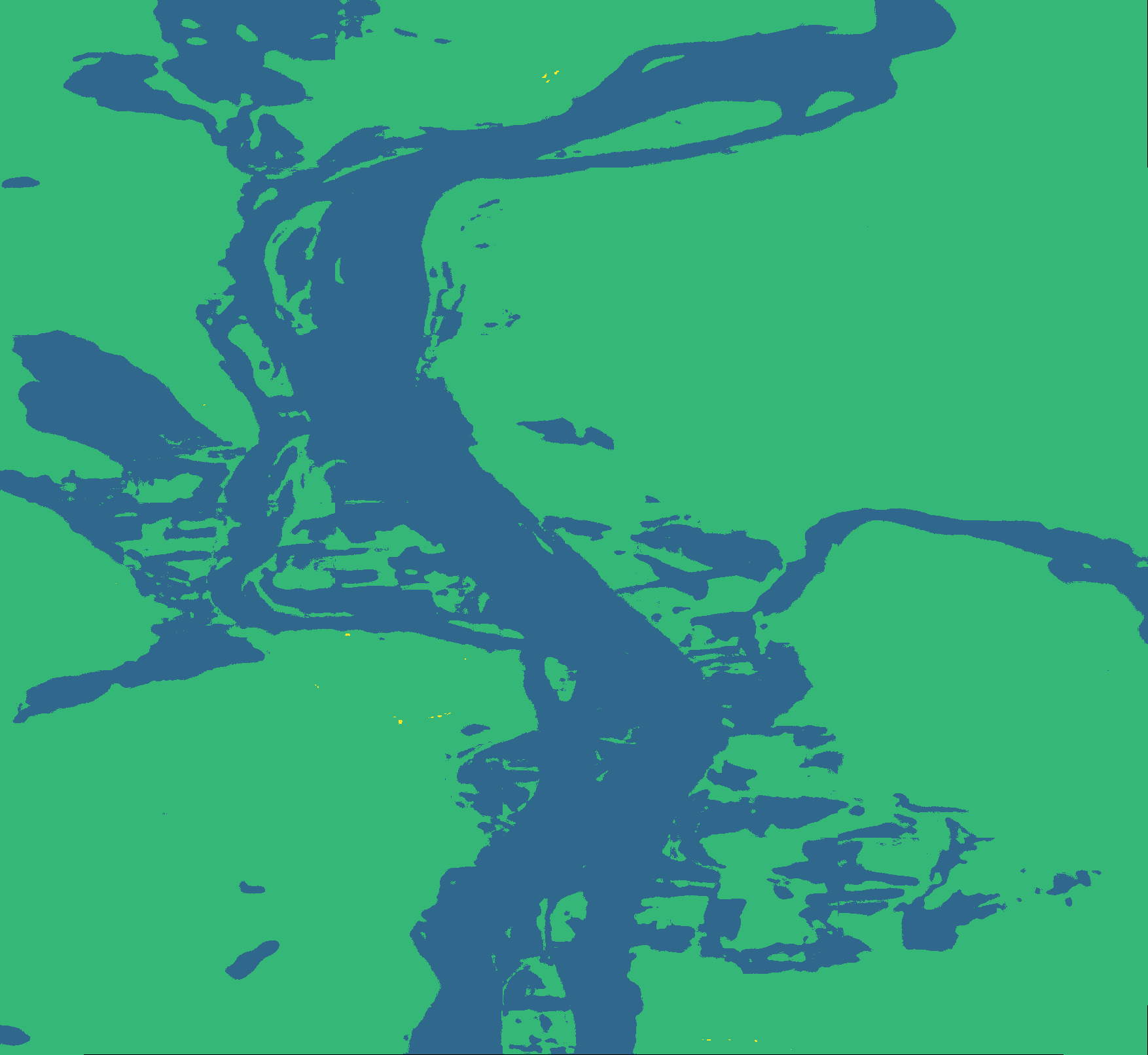}}\hfil
    \subfloat[]{\includegraphics[width=0.15\textwidth]{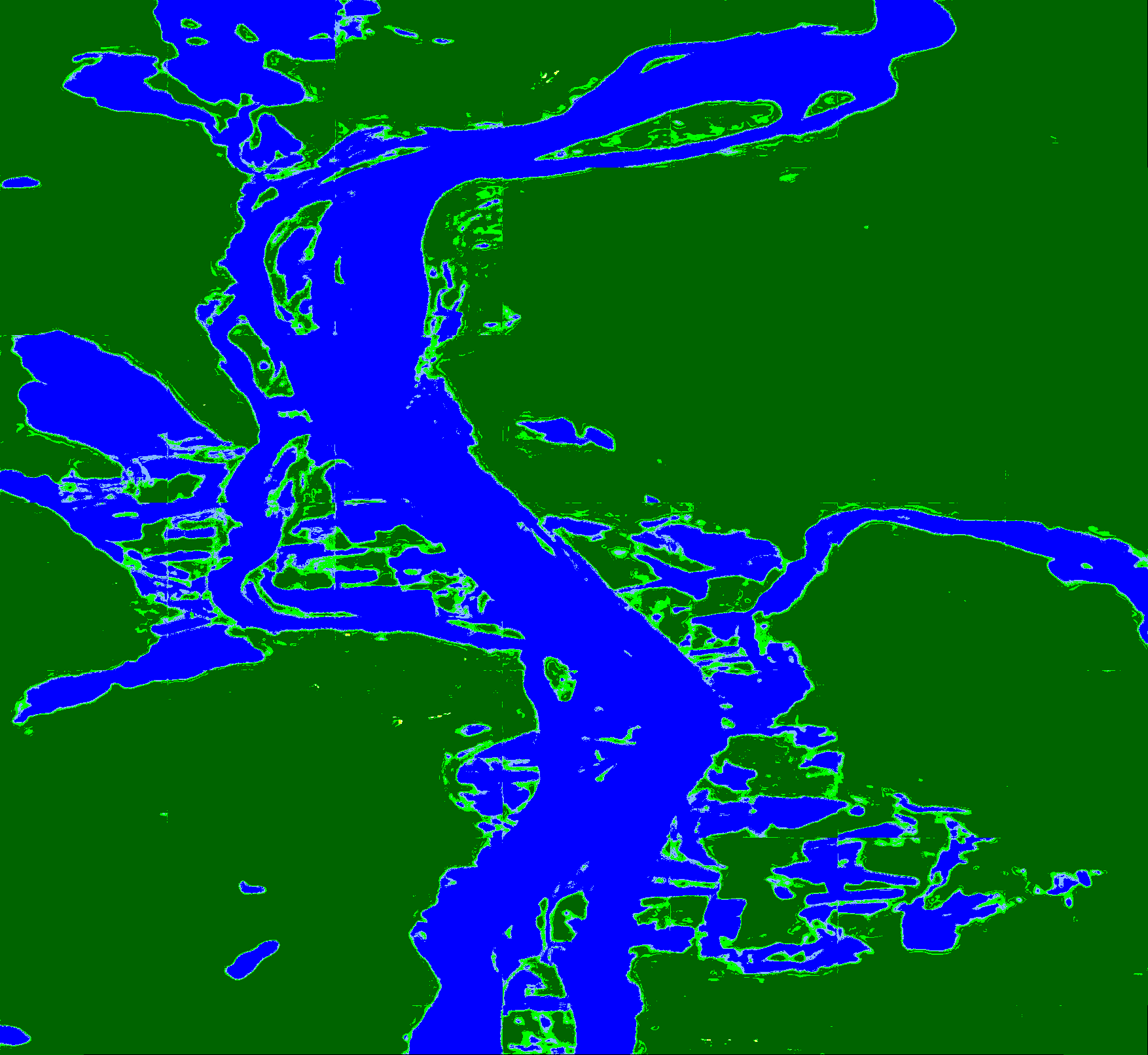}}\hfil
    
    \medskip
    \subfloat[]{\includegraphics[width=0.15\textwidth]{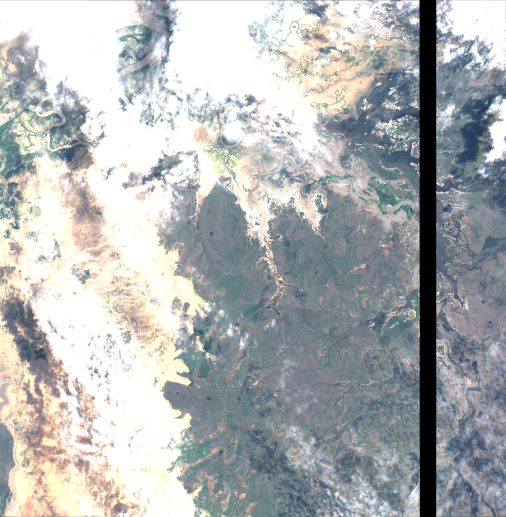}}\hfil
    \subfloat[]{\includegraphics[width=0.15\textwidth]{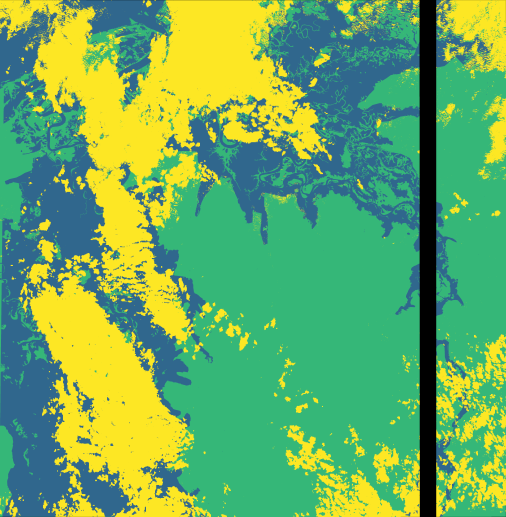}}\hfil
    \subfloat[]{\includegraphics[width=0.15\textwidth]{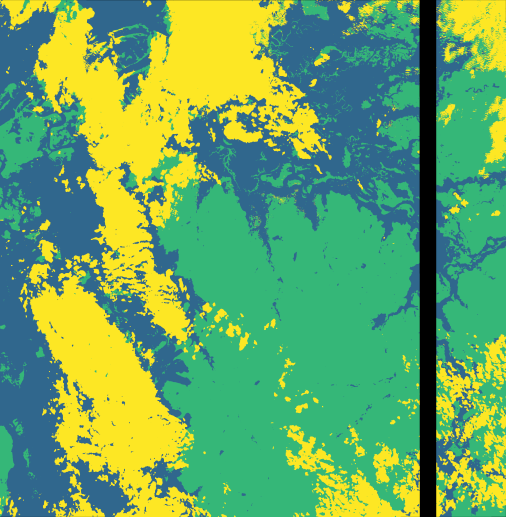}}\hfil
    \subfloat[]{\includegraphics[width=0.15\textwidth]{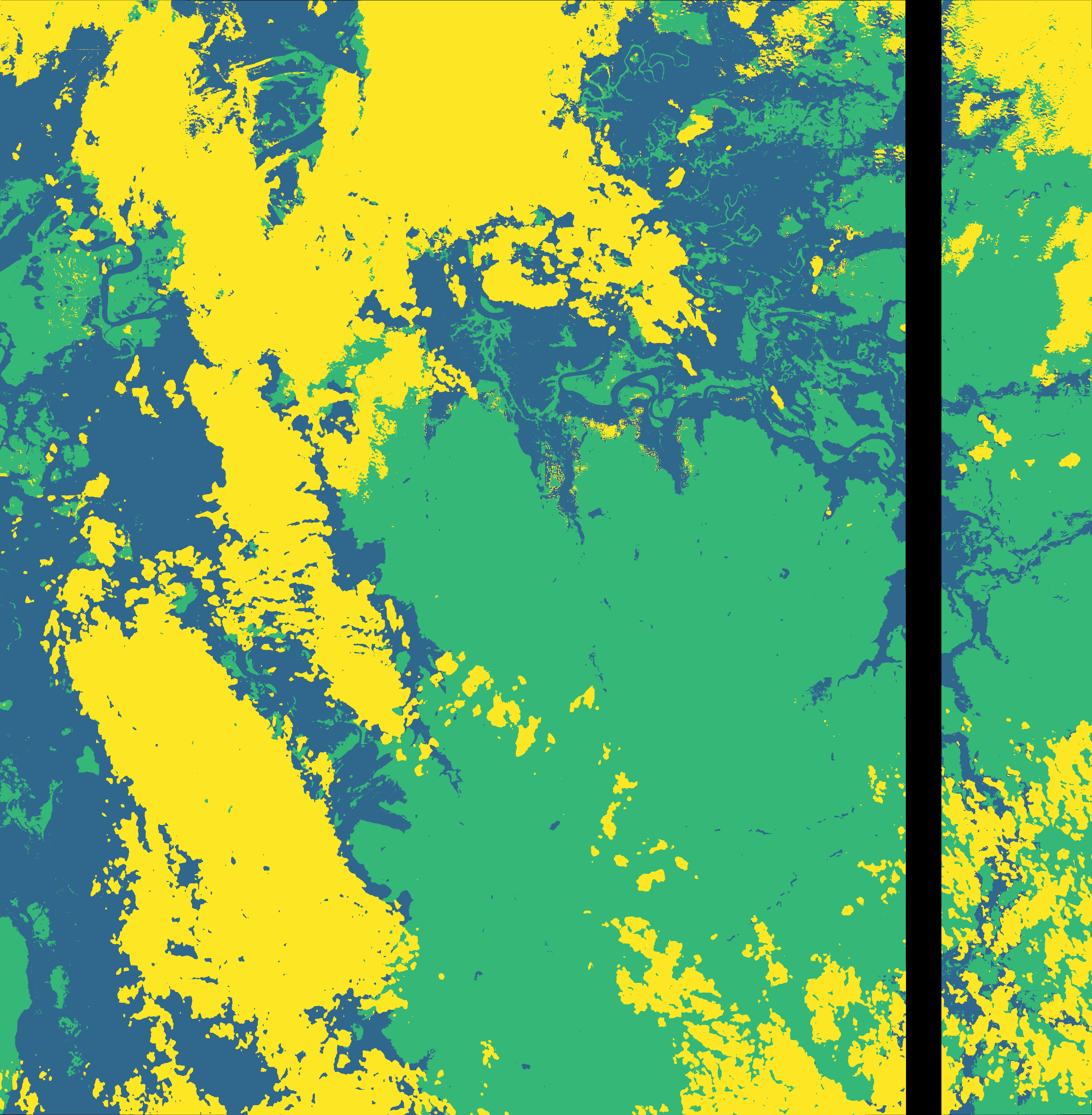}}\hfil
    \subfloat[]{\includegraphics[width=0.15\textwidth]{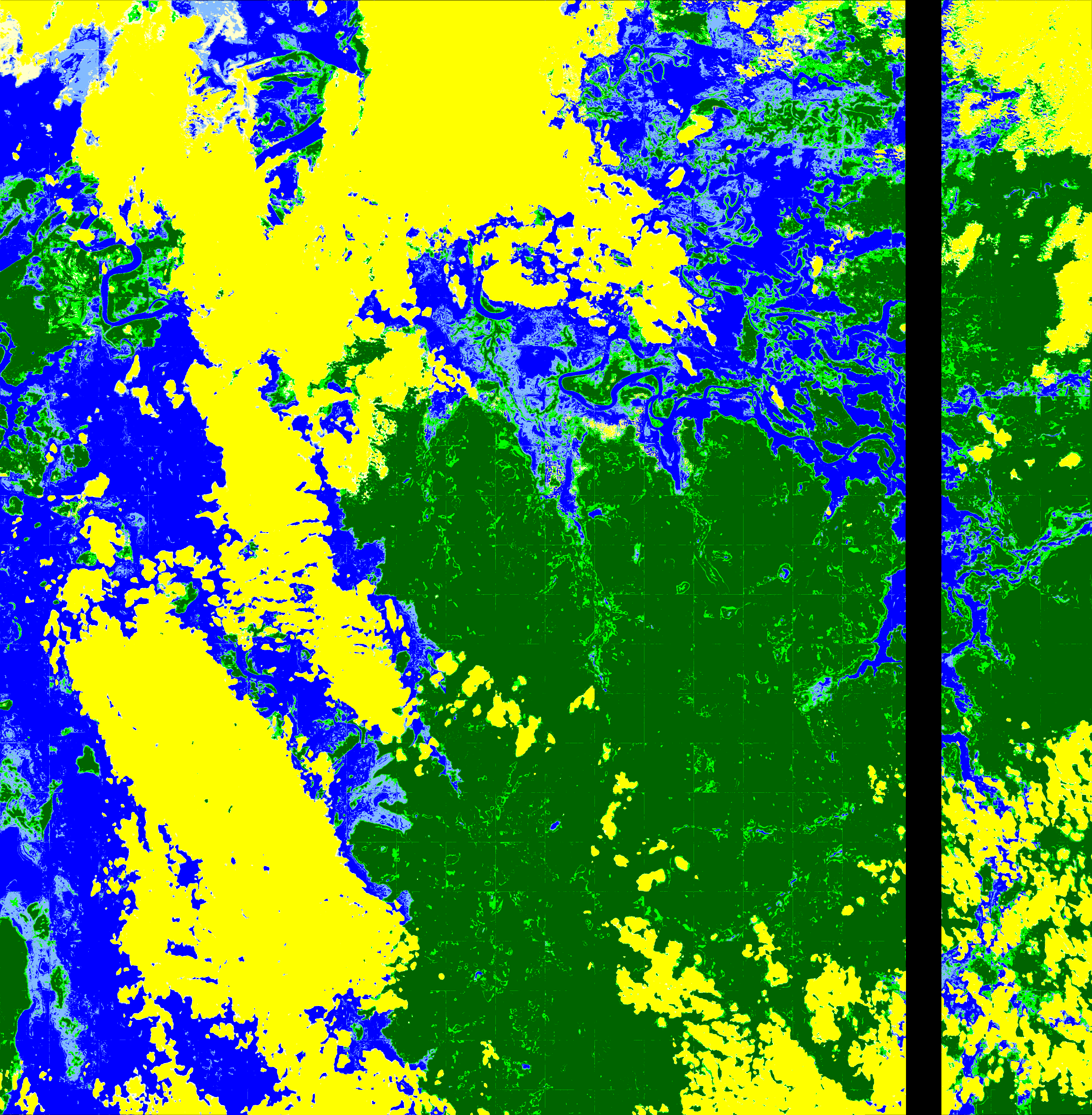}}\hfil

    \medskip
    \subfloat[RGB]{\includegraphics[width=0.15\textwidth]{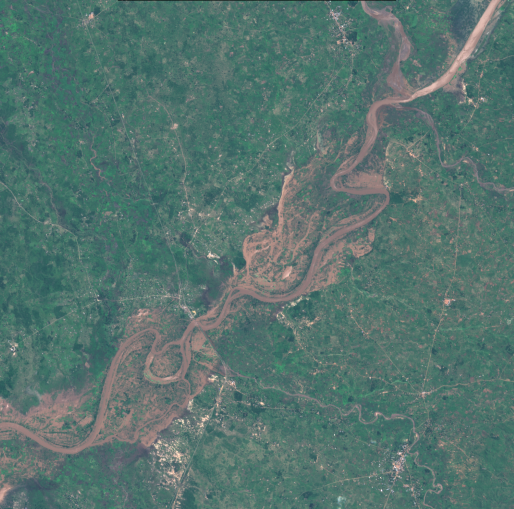}}\hfil
    \subfloat[Labels]{\includegraphics[width=0.15\textwidth]{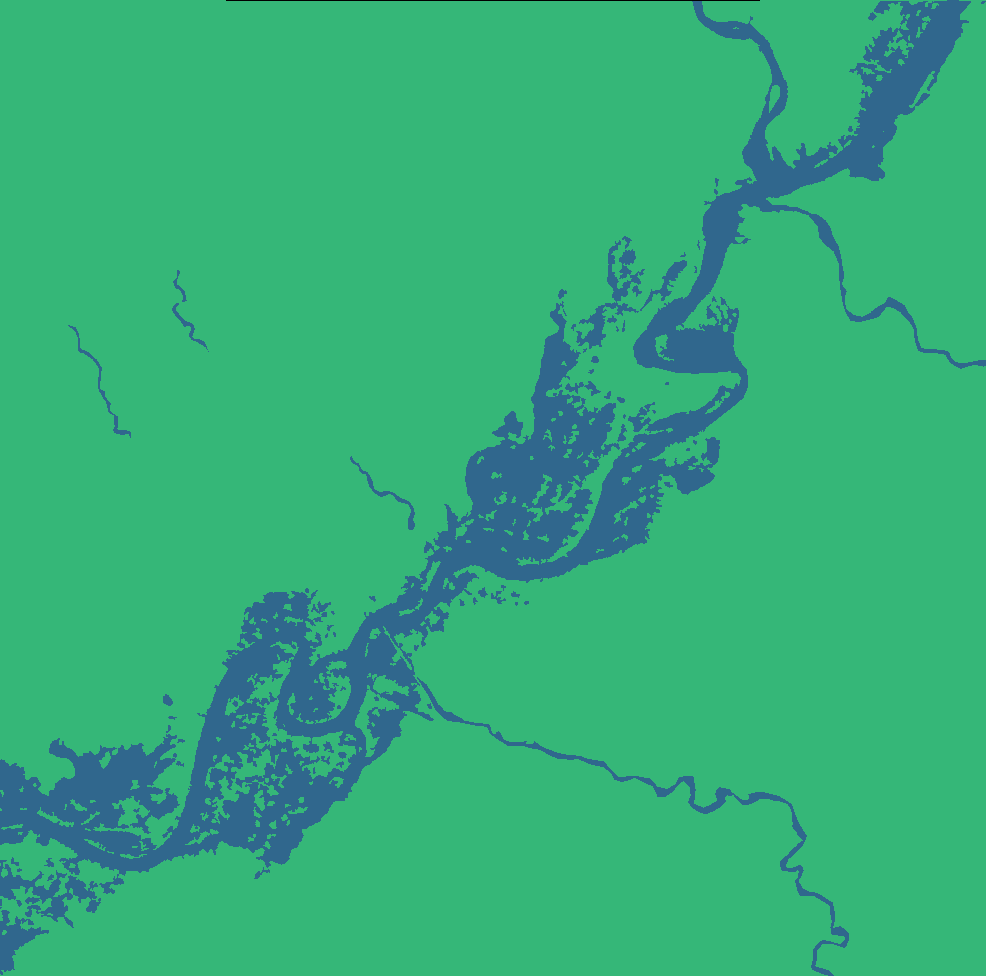}}\hfil
    \subfloat[U-Net]{\includegraphics[width=0.15\textwidth]{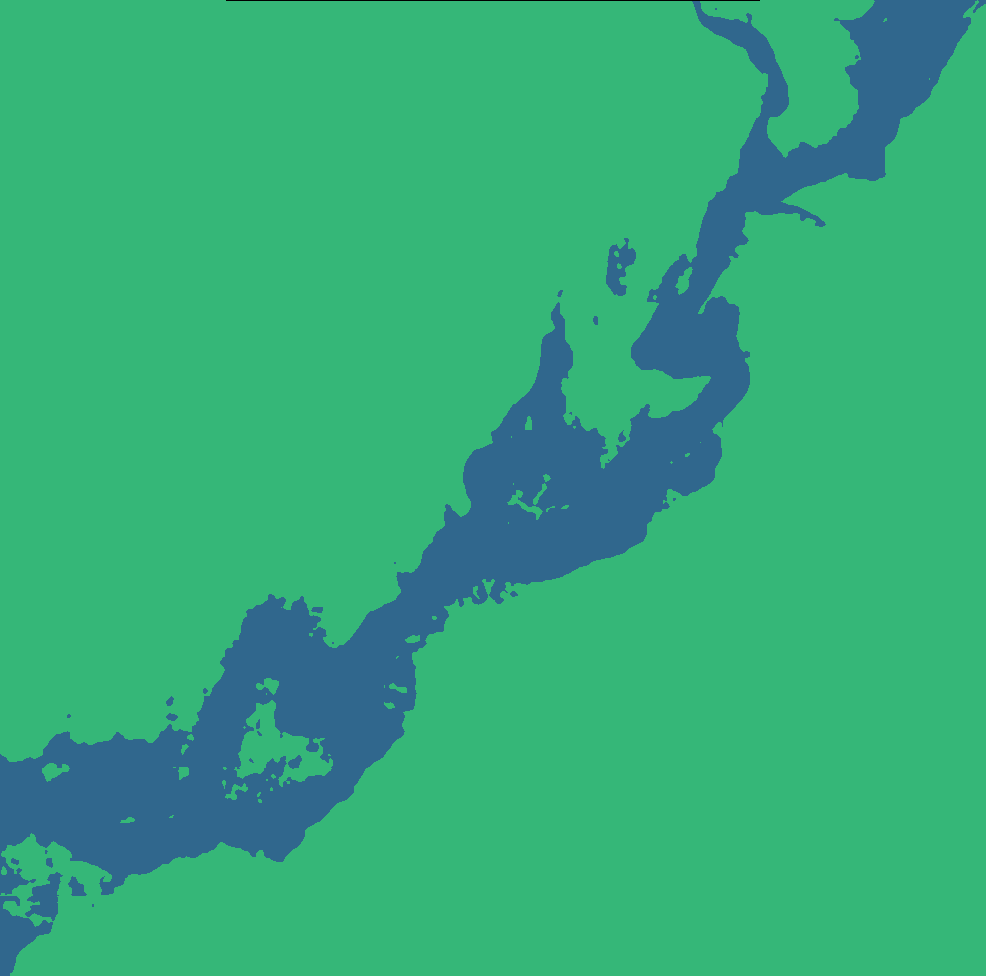}}\hfil
    \subfloat[IDSS+]{\includegraphics[width=0.15\textwidth]{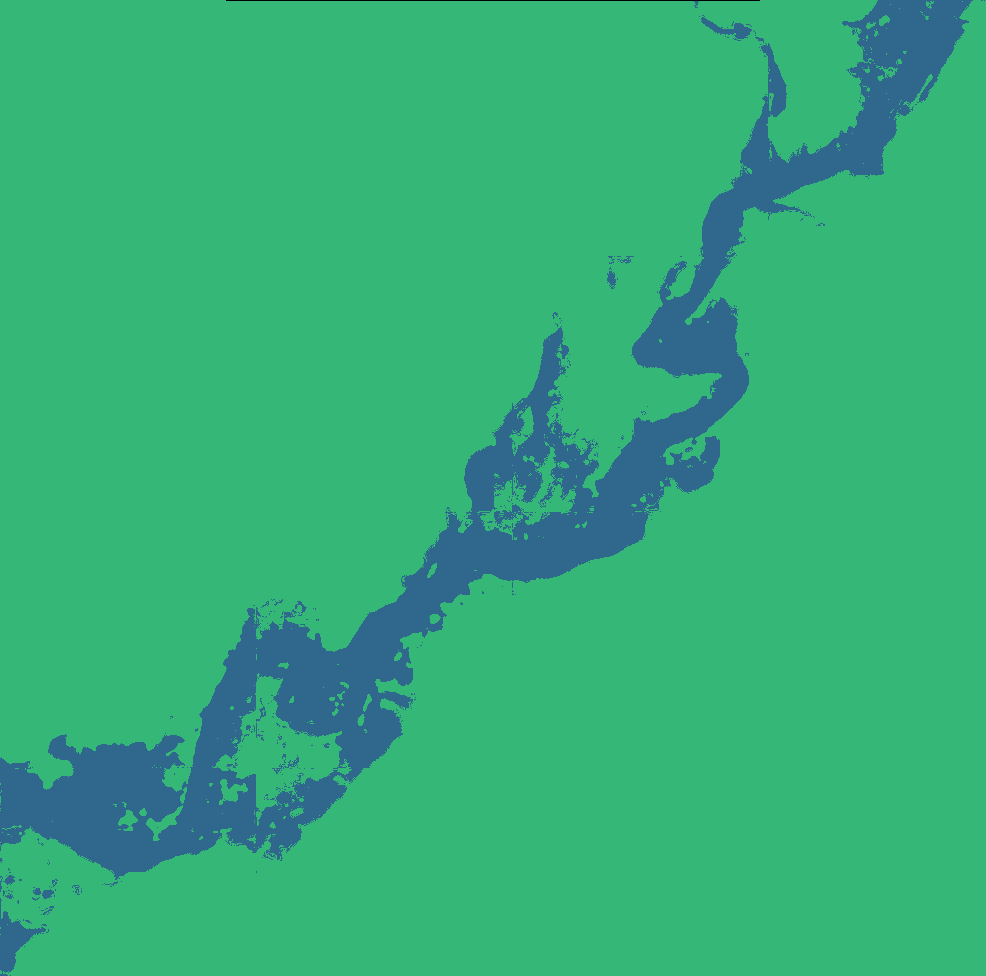}}\hfil
    \subfloat[Confidence map]{\includegraphics[width=0.15\textwidth]{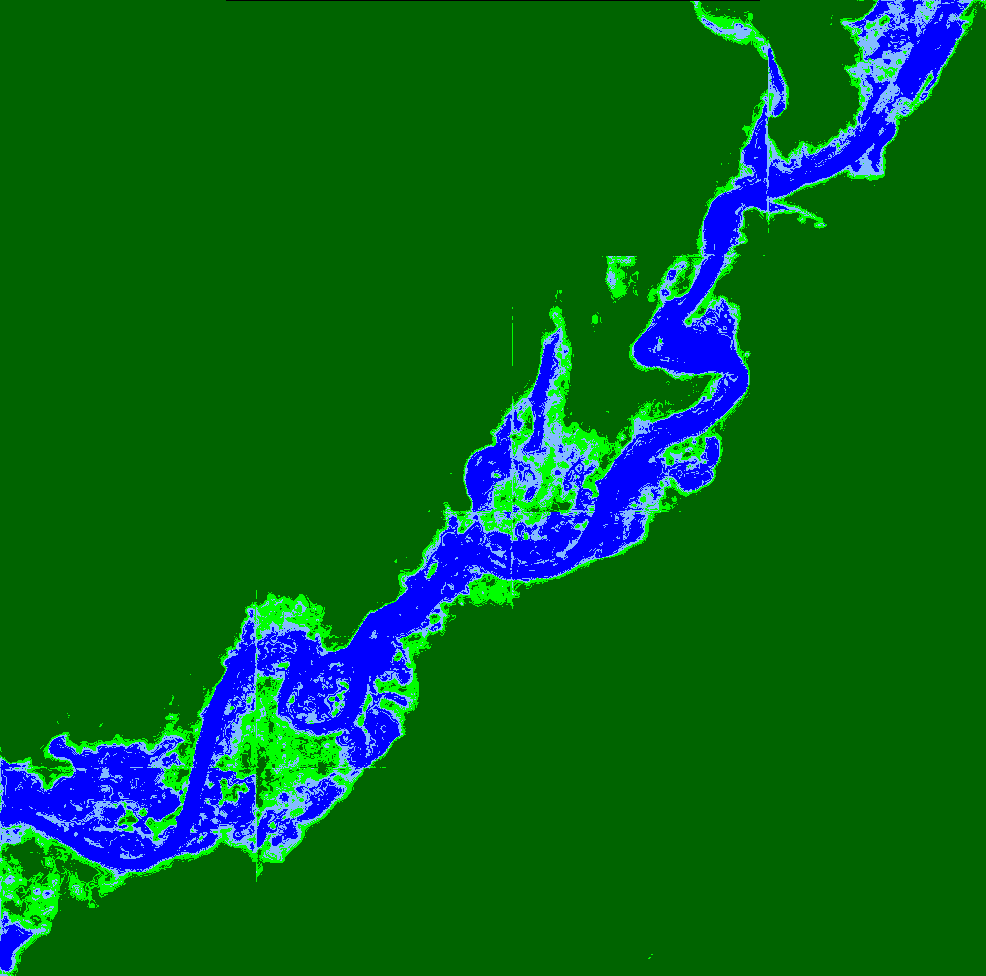}}\hfil

    \caption{Comparison of segmentation results. The meaning of the colours is: green - Land, yellow - Cloud, blue - Water. For the confidence map, lighter colors indicate lower confidence, while darker colors indicate high confidence.}
    \label{results}
\end{figure*}

\begin{figure*}[htbp]
    \centering
    \captionsetup[subfloat]{labelformat=empty}
    
    \subfloat[]{\includegraphics[width=0.15\textwidth]{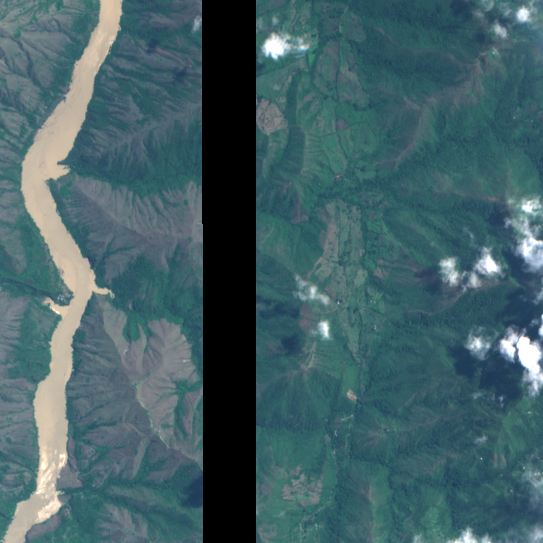}}\hfil
    \subfloat[]{\includegraphics[width=0.15\textwidth]{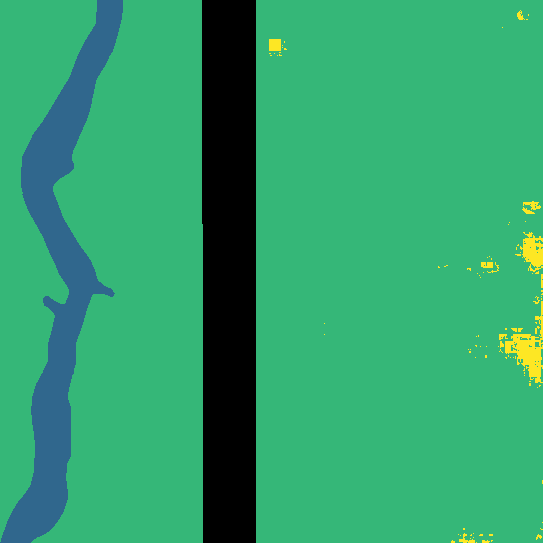}}\hfil
    \subfloat[]{\includegraphics[width=0.15\textwidth]{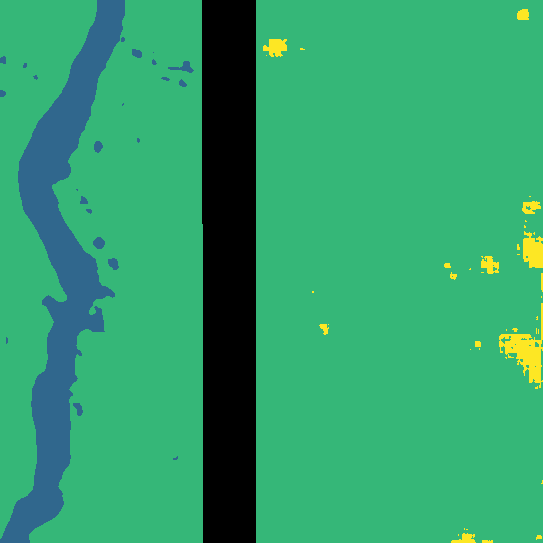}}\hfil
    \subfloat[]{\includegraphics[width=0.15\textwidth]{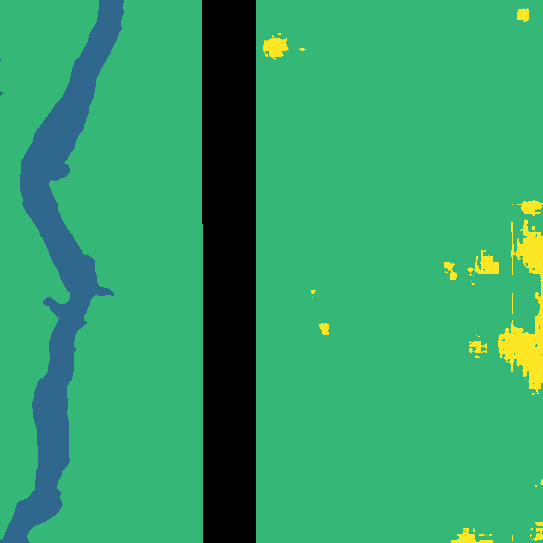}}\hfil

    \medskip
    \subfloat[RGB]{\includegraphics[width=0.15\textwidth]{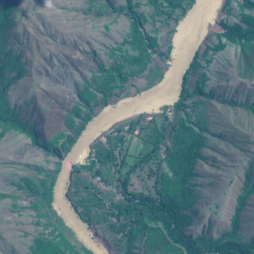}}\hfil
    \subfloat[Labels]{\includegraphics[width=0.15\textwidth]{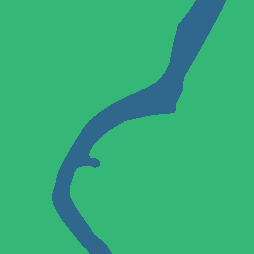}}\hfil
    \subfloat[U-Net]{\includegraphics[width=0.15\textwidth]{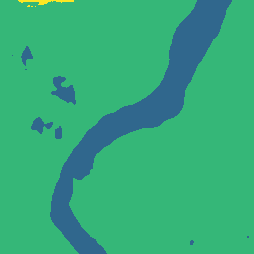}}\hfil
    \subfloat[IDSS+]{\includegraphics[width=0.15\textwidth]{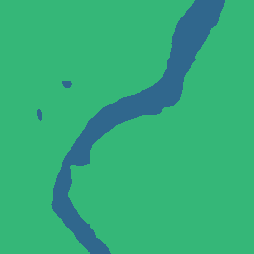}}\hfil

    \caption{Comparison of segmentation results. The meaning of the colours is: green - Land, yellow - Cloud, blue - Water.}
    \label{noise}
\end{figure*}

Since the pixel-wise similarity, $ s(x_{h,v}^k)$ itself measures the distance between the current pixel value and all previous pixel values at $(h,v)$th position, where $h$ denotes horizontal index within the images and $h = 1, 2, …, H$ means pixel index that varies between 1 and $H$ , $v$ denotes vertical index within the images and $v = 1, 2, …, V$ means pixel index that varies between 1 and $V$. The pixel-wise similarity, $s$, can naturally be used to perform binary change detection using a threshold, $\epsilon $:

\begin{equation}
    s(x_{h,v}^k) < \epsilon
\end{equation}

The derivation of the similarity, $s$, can be found in the Appendix.

\subsubsection{Semantic change detection}

In this paper, we propose the IDSS+ method and use it to label the binary change map obtained in the previous step. Both IDSS and IDSS+ use per-pixel prototypes. However, IDSS uses (averaged) centroid vectors, hence, it does not allow to associate the predictions with the real pixels from the training data. IDSS+ offers another option to make the decision-making process easier to understand for the users with a small loss of accuracy, which uses the nearest real pixels to each centroid as prototypes. 
This allows to present the segmentation problem using a dual form combining latent and raw feature space vectors.

The training and testing architecture of the IDSS+ is shown in Figure \ref{IDSS+}. During the training phase, each 13-band Sentinel-2 multispectral image $I \in \mathbb{R}^{H\times V \times n}$ is, first, fed into a feature extractor to obtain a latent feature vector $F \in \mathbb{R}^{H\times V \times D}$, where $n =13$ is the 13 bands of Sentinel-2 images and D represents the dimensionality of the latent feature space, e.g. for U-Net used in this paper, $D=64$. The feature extractor could be any semantic segmentation neural network. In this paper, the U-Net is used due to its excellent performance at water segmentation \citep{mateo2021towards}. It is important to note that U-Net has been trained on the Worldlfoods training dataset to achieve a better capability in feature extraction.

After feature extraction, for each pixel at position $(h,v)$, latent features $f_c^{h,v}$ and corresponding raw features $x_c^{h,v}$ were obtained within each class, $c \in{1,2,…C}$. For the Wordldfloods dataset, $C = 3$ since there are three classes in total, Water, Land, and Cloud. A clustering algorithm is then applied to the latent features to obtain the cluster centroids  $y_c^l$, where $y_c^l$ is the $l$th cluster center of class $c$, $l = 1, ..., L$ where $L$ is the number of clusters per class. After clustering the prototypes $P_c^l$, for which both the latent features that are closest to the clustering centroid and its corresponding raw features are selected. For clustering, the mini-batch k-means \citep{macqueen1967some, sculley2010web} or k-medoids \citep{kaufman1990partitioning} algorithms are selected and implemented. From Table \ref{tab:IoU comparison}, it can be observed that mini-batch k-means has better performance compared to k-medoids, but the k-medoids method could provide better interpretability because it uses real data observations as prototypes. An ablation study considering different numbers of prototypes for mini-batch k-means is shown in Table \ref{tab:K ablation study}.

The testing phase is also shown in Figure \ref{IDSS+}. The same feature extractor is used to obtain the latent features, $f^{h,v}$ from raw features, $x^{h,v}$. Then, the Euclidean distance is calculated between each latent feature $f^{h,v}$ of a given raw feature $x^{h,v}$ and the latent features $y_c^l$ from the prototypes $P_c^l$. Then, the k-nearest neighbour algorithm is used to make the final prediction.

Furthermore, a confidence map is introduced within IDSS+ to provide more insight into the decision-making process of the algorithm, as shown in Figure \ref{results}. Since the k-nearest neighbour algorithm is used to make the final decision, additional insight into the rationale and the strength/"confidence" of the decision can be obtained. If the "confidence" is plotted per pixel, we can obtain the heatmap shown as  Figure \ref{results}. A threshold is utilized to distinguish between high and low confidence, which can be adjusted by human users. In the context of Figure \ref{results}, the threshold is set to 0.8, which means that if more than 8 out of 10 neighbouring prototypes have the same class label, the decision is considered to be with high confidence. Predictions with lower confidence are indicated by lighter colors, while darker colors indicate high confidence.

An advantage of IDSS+ is its ability to significantly reduce noise pixels compared to U-Net as shown in Figure  \ref{noise}. It could be observed that the prediction of U-Net has much more noise pixels. On the contrary, IDSS+ provides a much cleaner and more accurate prediction. This is considered as the main reason that IDSS+ outperforms U-Net in terms of IoU water and Mean IoU, see Table \ref{tab:IoU comparison}.

In addition, IDSS+ provides more insight for human users by analyzing the prototypes which not only contain latent features, but also raw features.Furthermore, when k-medoid methods was used to select the prototypes, the prototypes are real training data. In this case, the human user can realistically observe the geographic location of the training data that is closest to the test data in the latent feature space and the spectral information contained in the corresponding raw features. For example, the new test pixel might come from the NorthWest of the UK and be predicted as a Water pixel, while the closest prototype is a Water pixel coming from the SouthEast of the UK. Such information could help the expert to further analyze the relationship between the current flood and historical information.

\subsubsection{Decision making}

This stage is in charge of combining the information obtained from the previous steps to perform the decision-making. Specifically, it provides the time when the flooding starts based on the percentage of change in the water pixels. After obtaining the semantic change map, the percentage of the change of the water pixels is calculated. Furthermore, the decision about “Flooding ” or “No Flooding” is made by setting a threshold that could be adjusted by human users. In addition, the previously obtained semantic change map can also provide human users with the exact location of the flooding.

\section{Experiments}

% \begin{figure}
%     \centering
%     \includegraphics[width=0.9\linewidth]{TP_actual_water.png}
%     \caption{High confidence example.}
%     \label{high_confidence}
% \end{figure}

% \begin{figure}
%     \centering
%     \includegraphics[width=0.9\linewidth]{TP_actual_water_pre_land_.png}
%     \caption{Low confidence example.}
%     \label{low_confidence}
% \end{figure}

\subsection{Datasets}

\begin{table*}[htbp]
    \centering
    \caption{IoU results comparison with other models.}
    \begin{tabular}{|c|c|c|c|c|}\hline
         Model  & IoU water (\%) & IoU land (\%) & IoU cloud (\%) & mIoU (\%)  \\
         \hline\hline
         $NDWI^1$ \citep{gao1996ndwi} & $65.12$ &  - & - & -  \\
         $NDWI^2$ \citep{gao1996ndwi}  & $39.99$ & - & - & - \\
         Linear \citep{mateo2021towards} &  $64.87$ & - & - & - \\
         SCNN \citep{mateo2021towards} & $71.12$  & - & - & -\\
         U-Net \citep{mateo2021towards} & $72.95$ & $82.56$ & $89.67$ & $81.71$ \\
         IDSS+ (k-means) & $\bold{74.52}$ & $\bold{82.72}$ & $88.46$ & $\bold{81.90}$\\
         IDSS+ (k-medoids) & $74.50$ & $82.54$ & $88.10$ & $81.71$\\
         \hline\hline
    \end{tabular}
    
    \label{tab:IoU comparison}
\end{table*}

\begin{table*}[htbp]
    \centering
    \caption{Comparison of IoU results for different numbers of prototypes for each class.}
    \begin{tabular}{|c|c|c|c|c|}\hline
        \#Prototypes  & IoU water (\%) & IoU land (\%) & IoU cloud (\%) & mIoU (\%)  \\
         \hline\hline
         $10$  & $71.80$ & $69.06$  &  $60.59$& $67.15$  \\
         $20$   & $73.23$ & $\bold{83.03}$ & $88.05$ & $81.44$ \\
         $50$  & $74.03$  & $82.08$ & $86.61$ & $80.91$ \\
         $100$  & $\bold{74.52}$ & $82.72$ & $\bold{88.46}$ & $\bold{81.90}$\\
         $1000$  & $74.00$ & $81.78$ & $88.01$  & $81.26$ \\
         $2000$  & $71.67$ & $82.71$ & $88.21$ & $80.86$\\
         $3000$  & $70.77$ & $82.34$ & $88.00$ & $80.37$\\
         \hline\hline
    \end{tabular}
    
    \label{tab:K ablation study}
\end{table*}

1) WorldFloods: WorldFloods \citep{mateo2021towards} is a semantic segmentation dataset tailored for flood detection. All flood maps were acquired from Sentinel-2 satellites. Specifically, there are three classes, Water, Land and Cloud. It contains 119 real flood events and has 424 flood maps with 407 images used for training, 6 for validation and 11 for testing.

\begin{table*}[htbp]
    \centering
    \caption{Performance comparison of IMAFD anomalous image detection on RavAEn dataset.}
    \begin{tabular}{|c|ccc|ccc|}\hline
     & \multicolumn{3}{l|}{Raw features (13 bands)}  & \multicolumn{3}{l|}{DINO-ViT-S/16 \citep{wang2022ssl4eo}} \\ \hline\hline
Scenario & Precision & Recall & F1 & Precision & Recall & F1 \\ \hline
1     & $\mathbf{0.11}$ & $\mathbf{1.00}$ & $\mathbf{0.20}$ & $0$ & $0$ & $0$ \\ \hline
2   & $\mathbf{0.50}$ & $1.00$ & $\mathbf{0.67}$ & $0.12$ & $1.00$ & $0.22$ \\ \hline
3   & $\mathbf{0.33}$ & $1.00$ & $\mathbf{0.50}$ & $0.20$ & $1.00$ & $0.33$ \\ \hline
4  & $0.17$ & $1.00$ & $0.29$ & $0.17$ & $1.00$ & $0.29$ \\ \hline
Average  & $\mathbf{0.28}$ & $\mathbf{1.00}$ & $\mathbf{0.42}$ & $0.12$ & $0.75$ & $0.21$ \\ \hline
\end{tabular}
    
    \label{RavAen_results}
\end{table*}

\begin{figure*}
  \centering

  \medskip

  \begin{subfigure}[t]{.4\linewidth}
    \centering\includegraphics[width=1.0\linewidth]{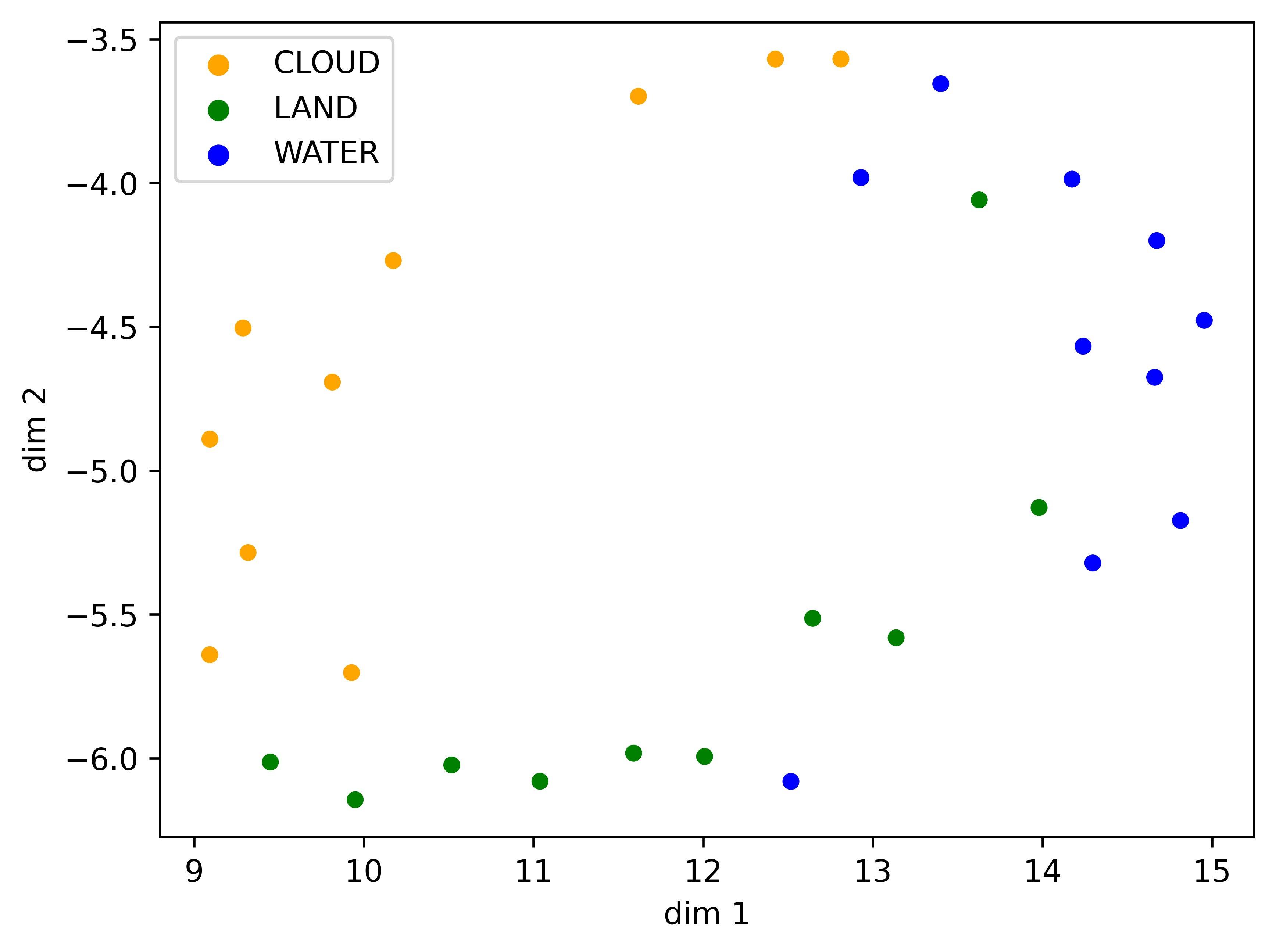}
    \caption{\#Prototypes for each class : 10}
    \label{correct_pre}
  \end{subfigure}\quad
  \begin{subfigure}[t]{.4\linewidth}
    \centering\includegraphics[width=1.0\linewidth]{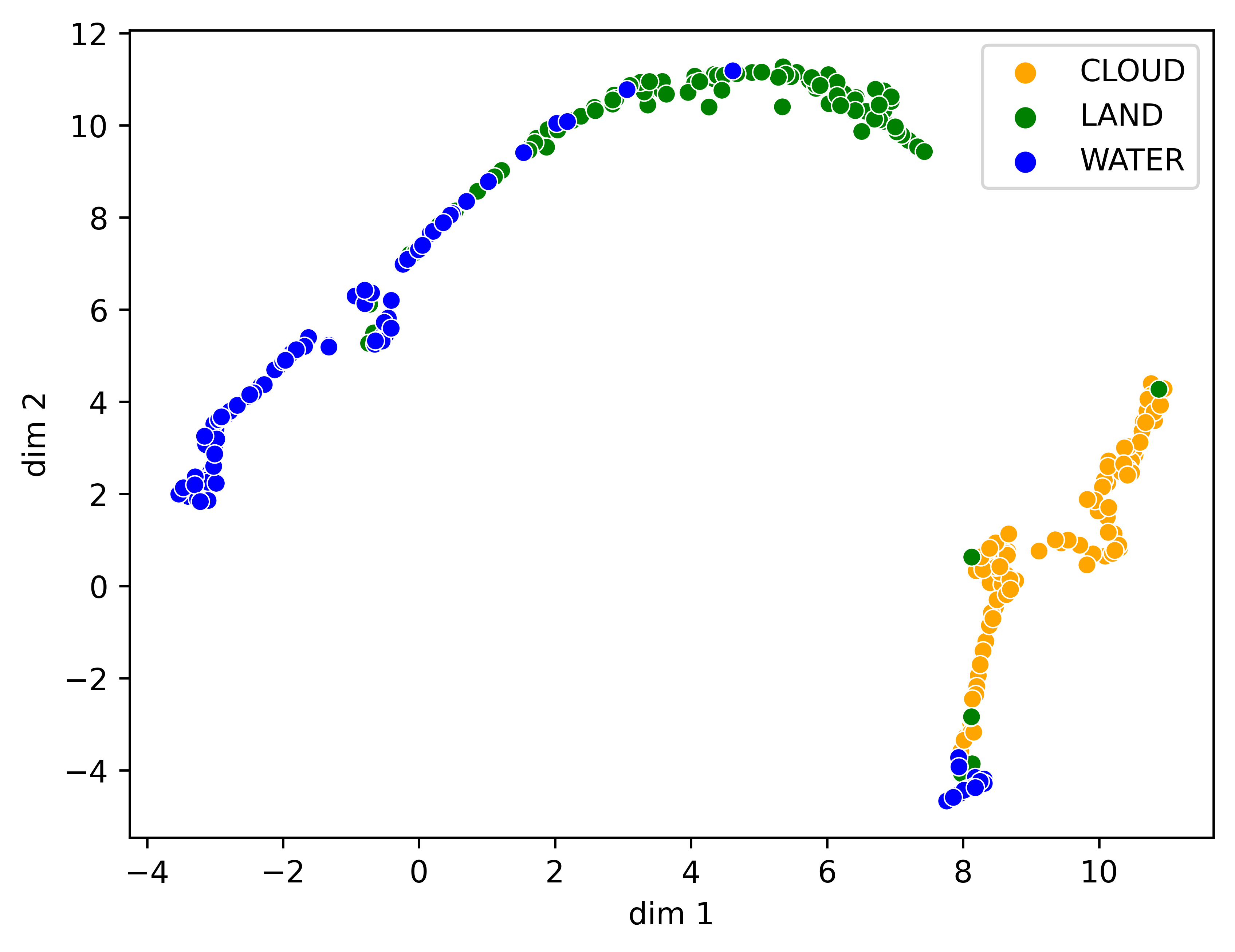}
    \caption{\#Prototypes for each class : 100}
    \label{incorrect_pre}
  \end{subfigure}

  \caption{Prototype visualisation in the latent space with different number of prototypes using Uniform Manifold Approximation and Projection (UMAP)}
  \label{UMAP}
\end{figure*}

\begin{figure*}[htbp]
  \centering

  \medskip

  \begin{subfigure}[t]{.4\linewidth}
    \centering\includegraphics[width=1.0\linewidth]{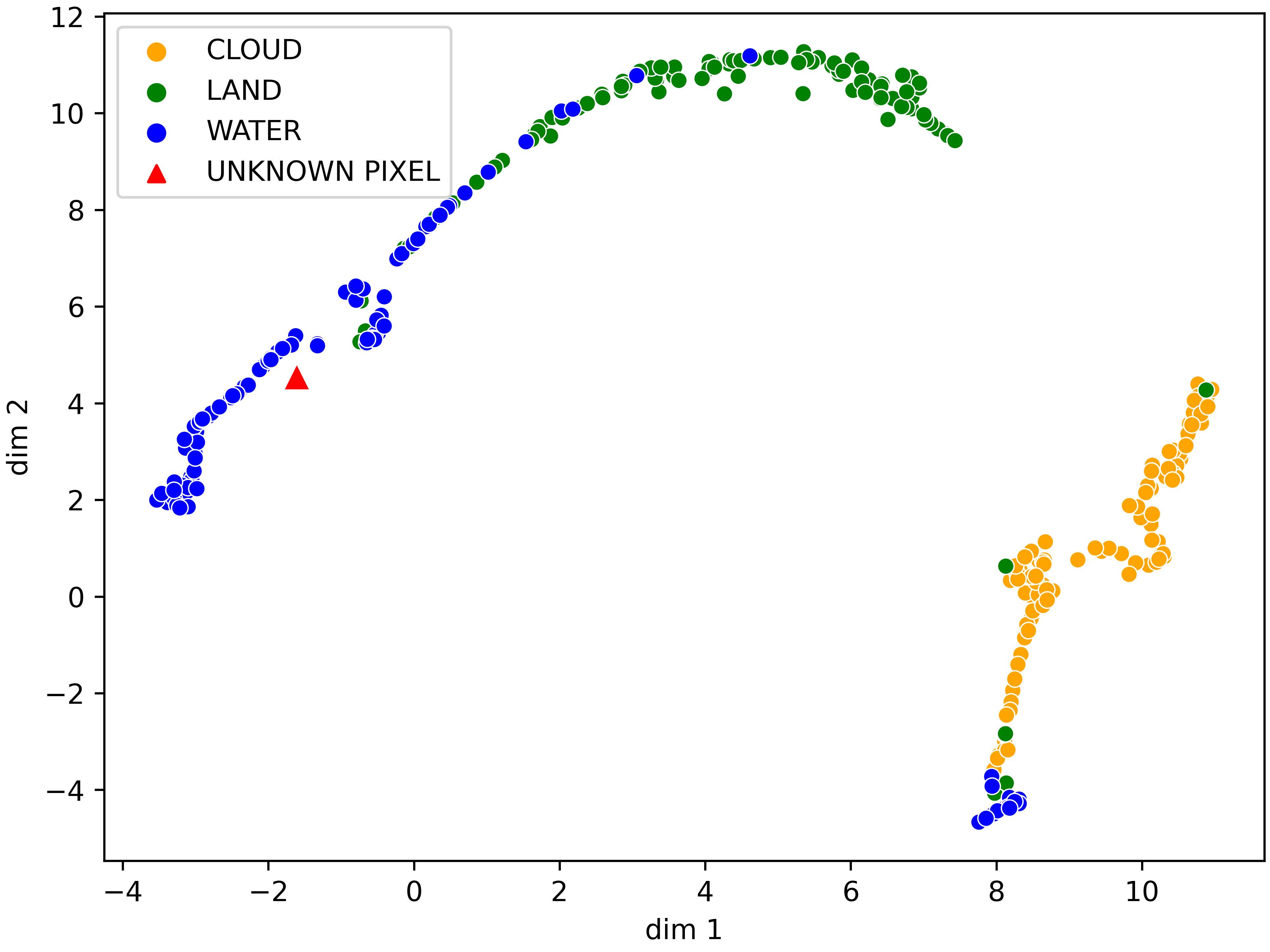}
    \caption{UMAP plot for a correct prediction (prediction: Water, label : Water)}
    \label{correct_pre}
  \end{subfigure}\quad
  \begin{subfigure}[t]{.4\linewidth}
    \centering\includegraphics[width=1.0\linewidth]{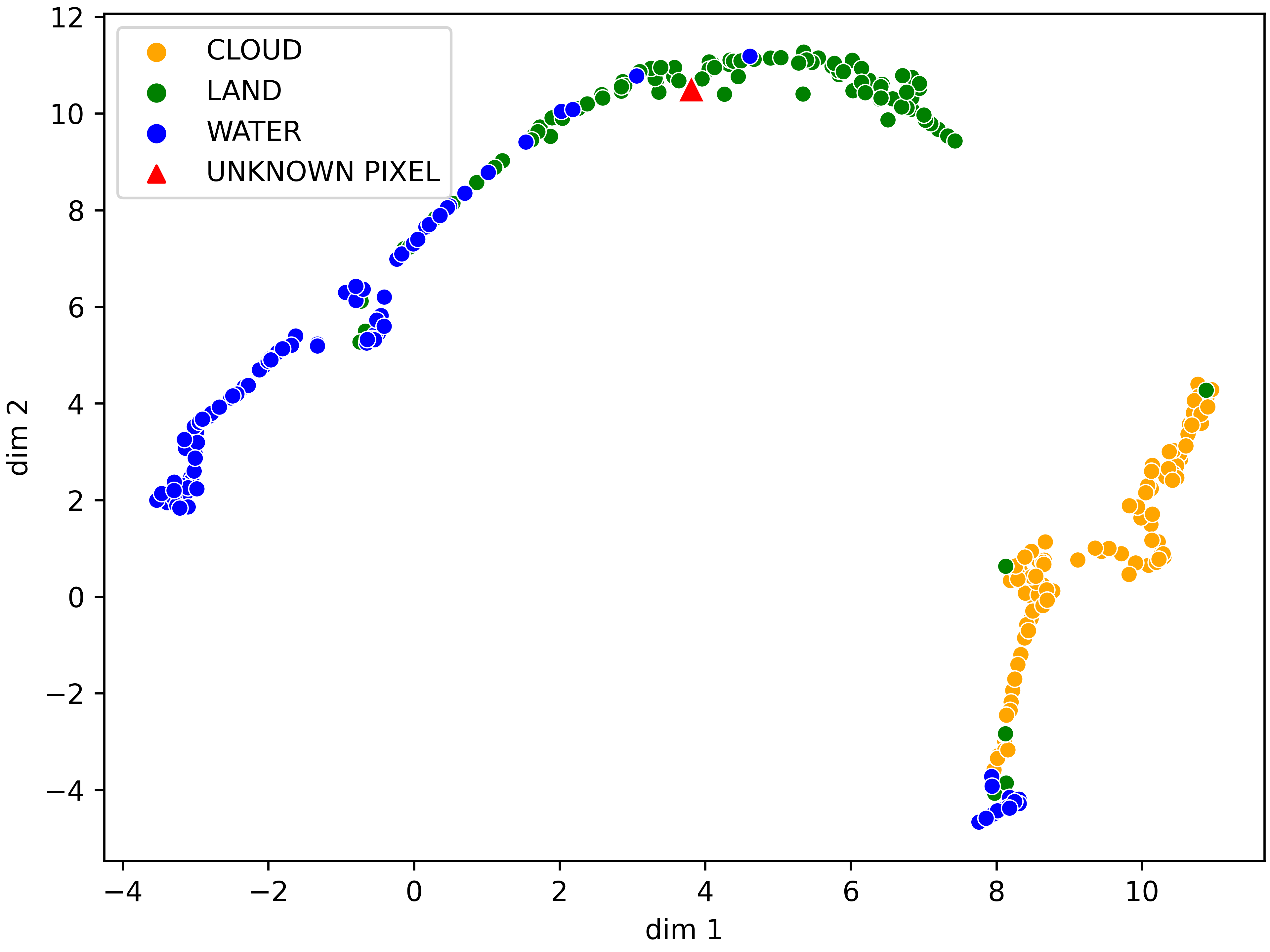}
    \caption{UMAP plot for a wrong prediction (prediction: Land, label: Water)}

    \label{incorrect_pre}
  \end{subfigure}

  \caption {Visualisation of the decision-making process with the prototype in the latent space using UMAP. Plot (a) shows an example where the unknown red triangular pixel is successfully predicted as Water because it is much closer to the blue Water prototypes, while plot (b) shows an example where the unknown red triangular pixel is misclassified as Land because it is surrounded by green Land prototypes, although its ground truth label is Water.}
  \label{decision_making}
\end{figure*}

% \begin{figure*}[htbp]
%     \centering
%     \captionsetup[subfloat]{labelformat=empty}

%     \medskip
%     \subfloat[10]{\includegraphics[width=0.14\textwidth]{Umap_10.png}}\hfil
%     \subfloat[20]{\includegraphics[width=0.14\textwidth]{Umap_20.png}}\hfil
%     \subfloat[50]{\includegraphics[width=0.14\textwidth]{Umap_50.png}}\hfil
%     \subfloat[100]{\includegraphics[width=0.14\textwidth]{Umap_100.png}}\hfil
%     \subfloat[1000]{\includegraphics[width=0.14\textwidth]{Umap_1000.png}}\hfil
%     \subfloat[2000]{\includegraphics[width=0.14\textwidth]{Umap_2000.png}}\hfil
%     \subfloat[3000]{\includegraphics[width=0.14\textwidth]{Umap_3000.png}}\hfil

%     \caption{Prototype visualisation in the latent space with different number of prototypes using Uniform Manifold Approximation and Projection (UMAP)}
%     \label{results}
% \end{figure*}

\begin{table}[htbp]
    \centering
    \caption{Performance evaluation of IMAFD anomalous image detection on MediaEval dataset.}
    \begin{tabular}{|c|c|c|c|c|}\hline
         Scenarios & Precision  & Recall & F1 score   \\
         
         \hline\hline

         1 & $0.82$ &$1.00$  & $0.90$ \\
         2 & $0.82$ & $1.00$ & $0.90$\\
         3 & $0.82$ & $1.00$ & $0.90$\\
         4 & $0.82$ & $1.00$ & $0.90$\\
         5 & $0.82$ & $1.00$ & $0.90$\\
         6 & $0.75$ & $1.00$ & $0.86$\\
         7 & $0.75$ & $1.00$ & $0.86$\\
         8 & $0.90$ & $1.00$ & $0.95$\\
         9 & $0.75$ & $1.00$ & $0.86$\\

         \hline\hline
        
    \end{tabular}

    \label{media_eval_results}
\end{table}

\begin{table*}[htpb]
\centering
\caption{Semantic segmentation change results comparison}
\begin{tabular}{|l|llll|llll|}

\hline
         & \multicolumn{4}{c|}{E-IMAFD (IDSS+)}              & \multicolumn{4}{c|}{E-IMAFD (U-Net)}                         \\ \hline\hline
Scenario & Precision & Recall & F1 & IoU water & Precision & Recall & F1 & IoU water \\ \hline
1        & $\mathbf{83.64}$     & $79.22$  & $\mathbf{81.37}$  & $68.59$   & $81.93$     & $\mathbf{79.39}$  & $80.64$  & $67.56$ \\ \hline
2        & $\mathbf{60.87}$     & $89.51$  & $\mathbf{72.46}$  & $\mathbf{56.82}$   & $57.79$     & $\mathbf{90.71}$  & $70.60$  & $54.56$ \\ \hline
3        & $\mathbf{43.51}$     & $91.13$  & $\mathbf{58.90}$  & $\mathbf{41.74}$   & $41.90$     & $\mathbf{91.66}$  & $57.50$  & $40.35$ \\ \hline
4        & $\mathbf{65.63}$     & $80.24$  & $72.20$  & $56.50$   & $64.16$     & $\mathbf{86.75}$  & $\mathbf{73.77}$  & $\mathbf{58.44}$ \\ \hline
Average  & $\mathbf{63.41}$     & $85.03$  & $\mathbf{71.23}$  & $\mathbf{55.91}$   & $61.45$     & $\mathbf{87.13}$  & $70.63$  & $55.23$ \\ \hline
\end{tabular}

\label{semantic_change_results_table}
\end{table*}

2) RaVAEn: The RaVAEn \citep{ruuvzivcka2022ravaen} dataset contains four different natural disasters and since our paper mainly focuses on floods, we consider four sets of time series images captured by Snetinel-2 satellites at different locations. We expanded the flood dataset so each set now contains 15 images instead of the 5 contained in the original dataset. 14 of the images were taken before the flood, and the last one was taken after the flood. The extended dataset was downloaded using the Google Earth Engine and is also Sentinel-2A images, ensuring that it is the same size and location as the original dataset. Each post-flood image provides a ground truth label manually annotated by human experts.

3) MediaEval: The MediaEval dataset \citep{bischke2019multimedia} contains 335 sets of Sentinel-2 time series images.  Each set contains between 4 and 20 images, taken either 45 days before or 45 days after the flood. Each image was labeled as either “Flooding” or “Non-flooding” by experts from Copernicus Emergency Management Service (EMS). To demonstrate the efficiency of the first phase of IMAFD, 10 sets of images were selected.

\subsection{Evaluation metrics}

The evaluation metrics can be divided into three categories. The first one mainly evaluates the performance of the first stage of the IMAFD for anomaly detection of Sentinel-2 time series images. Precision, Recall and F1 scores were calculated and compared. The second one is to evaluate the semantic segmentation performance of the IDSS+ on the WorldFloods dataset, Intersection over Union (IoU) is mainly used, including IoU water, IoU land, IoU cloud, and Mean IoU (mIoU). The latter is used to evaluate the semantic change map obtained after the third stage of semantic change detection. Precision, Recall, F1 score and IoU water are utilized.

\subsection{Results}

1)	Anomalous image detection

To evaluate the performance of the first stage of IMAFD (anomalous image detection), experiments were conducted on both the RavAEn and MediaEval datasets. The results are presented in Tables \ref{RavAen_results} and \ref{media_eval_results}.  The RSE method used in the anomalous image detection stage is sensitive to the input features. We performed an ablation study to compare its performance under different features on the RavAEn dataset, as shown in Table \ref{RavAen_results}. It can be observed that when using the raw features, the performance is better than the case when latent features derived from the DINO-ViT-S/16 model pre-trained on EO datasets \citep{ wang2022ssl4eo} were used. Specifically, all the recall values obtained by raw features are 1, meaning that no flood event was missed. On the MediaEval dataset, we only tested the raw features and still got all recall values as 1. 

\begin{figure*}[htbp]
    \centering
    \captionsetup[subfloat]{labelformat=empty}
    
    \subfloat[]{\includegraphics[width=0.15\textwidth]{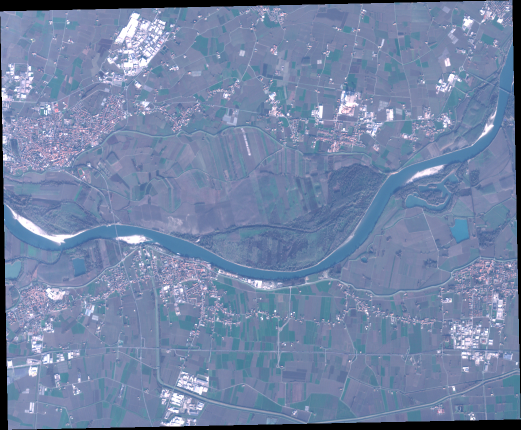}}\hfil
    \subfloat[]{\includegraphics[width=0.15\textwidth]{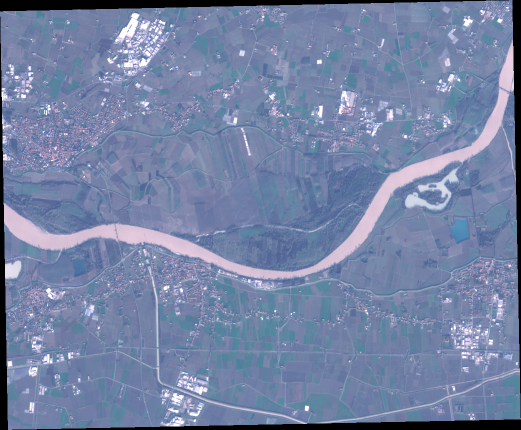}}\hfil
    \subfloat[]{\includegraphics[width=0.15\textwidth]{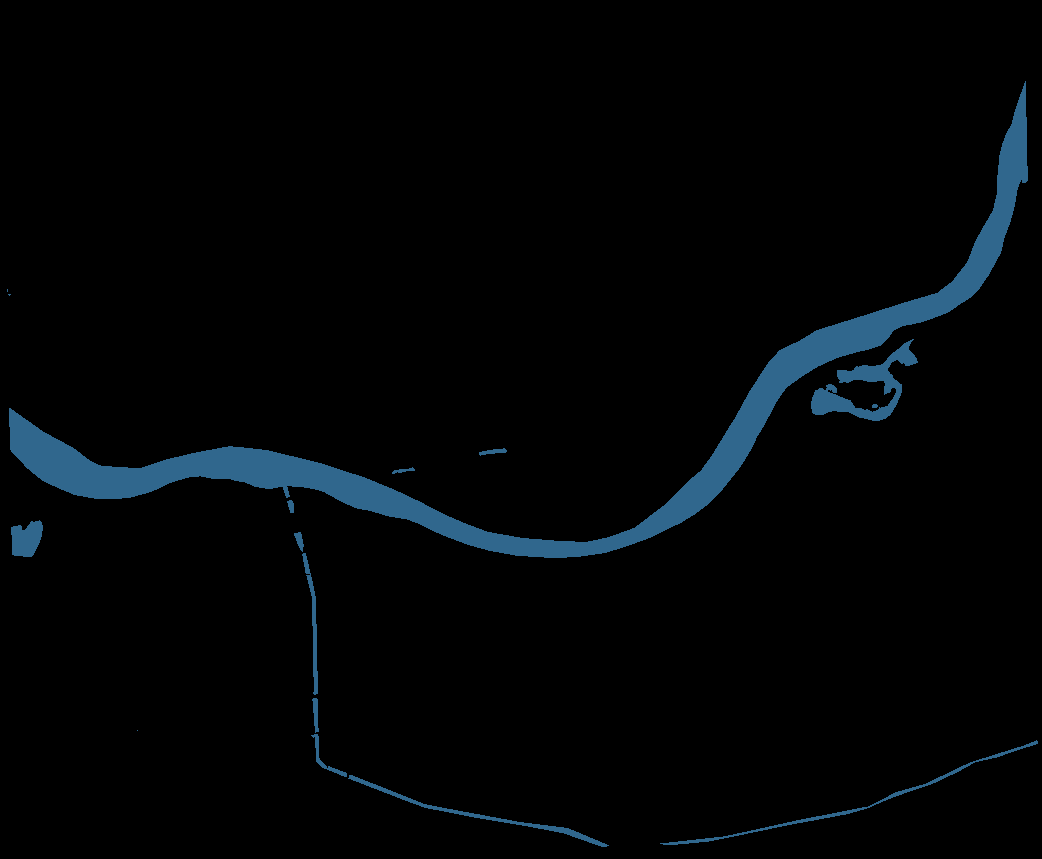}}\hfil
    \subfloat[]{\includegraphics[width=0.15\textwidth]{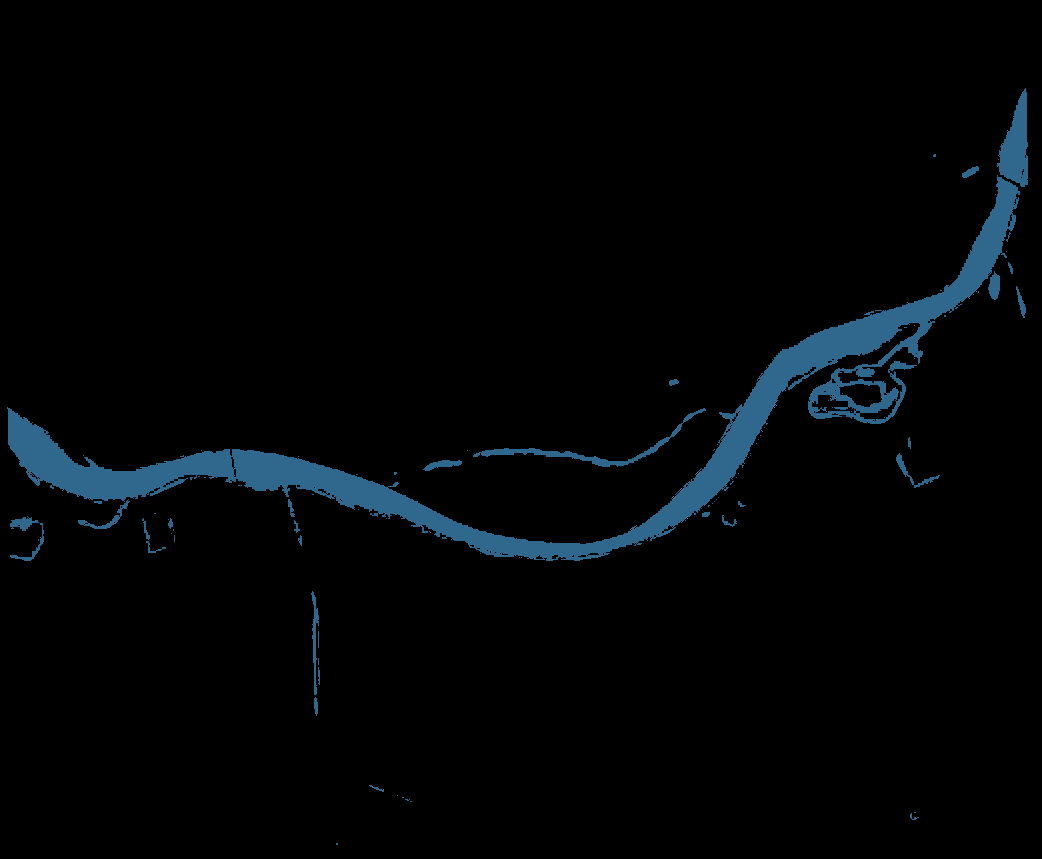}}\hfil
    \subfloat[]{\includegraphics[width=0.15\textwidth]{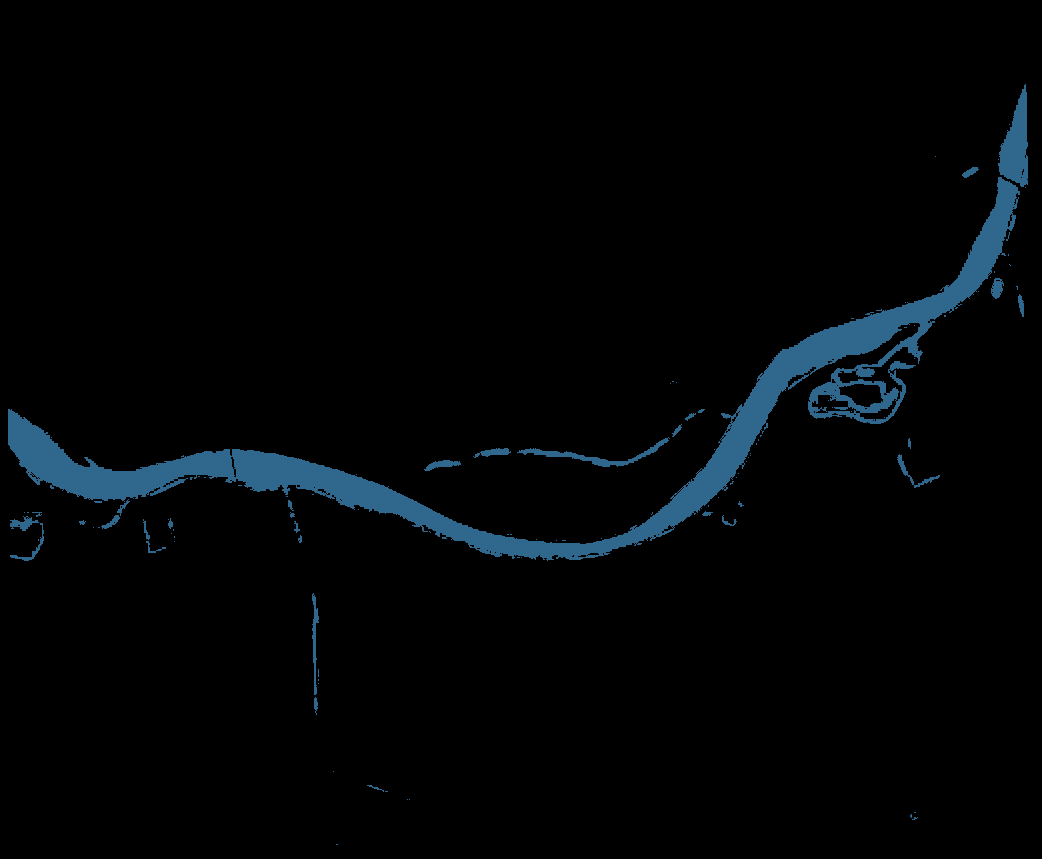}}\hfil
    
    \medskip
    \subfloat[]{\includegraphics[width=0.15\textwidth]{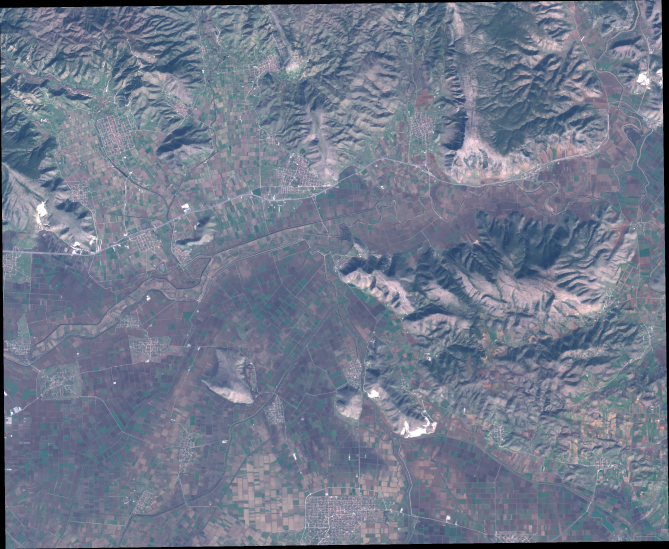}}\hfil
    \subfloat[]{\includegraphics[width=0.15\textwidth]{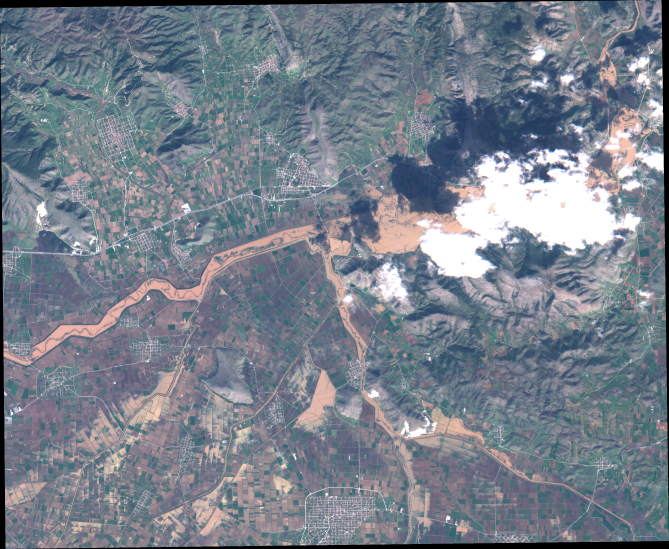}}\hfil
    \subfloat[]{\includegraphics[width=0.15\textwidth]{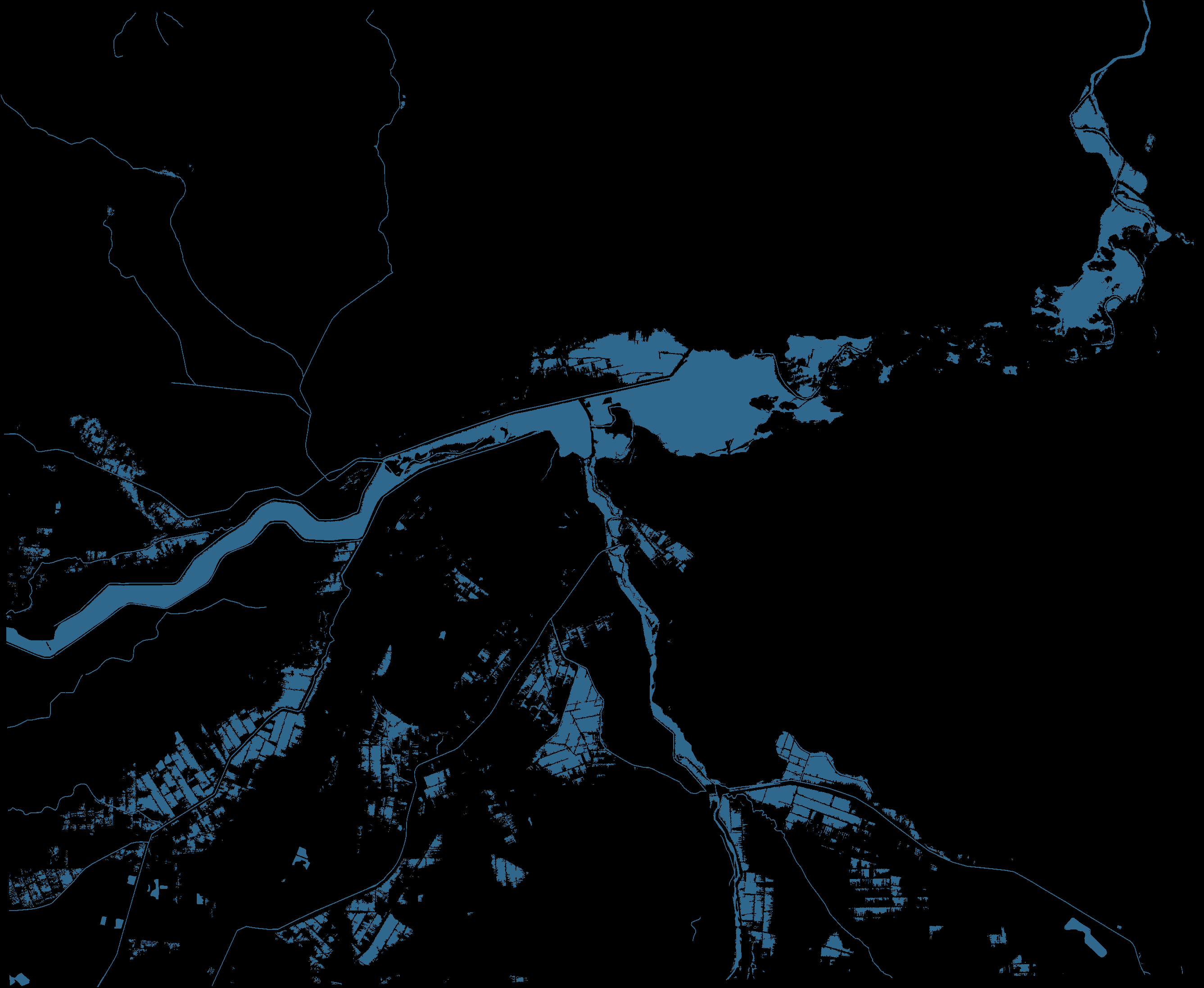}}\hfil
    \subfloat[]{\includegraphics[width=0.15\textwidth]{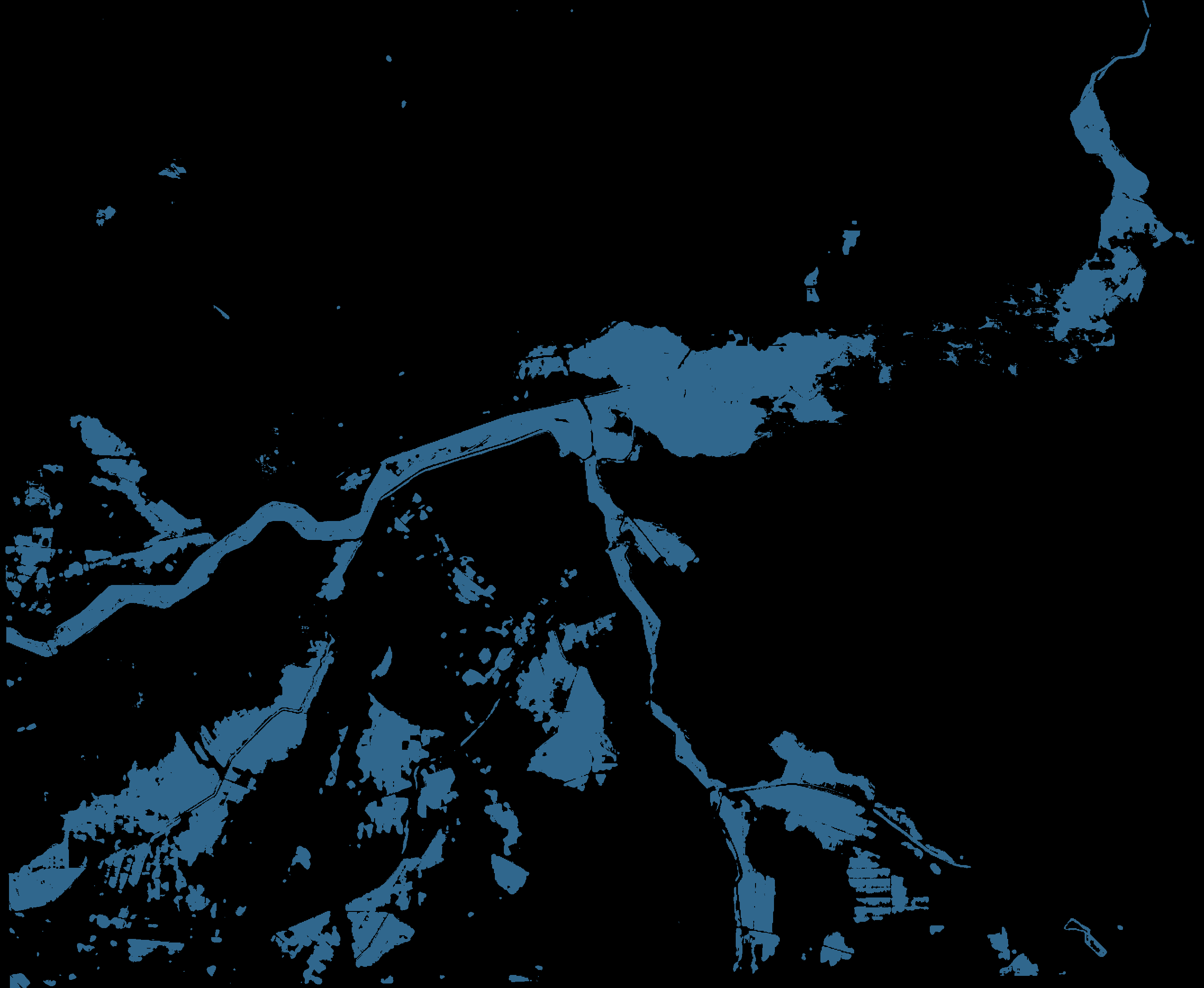}}\hfil
    \subfloat[]{\includegraphics[width=0.15\textwidth]{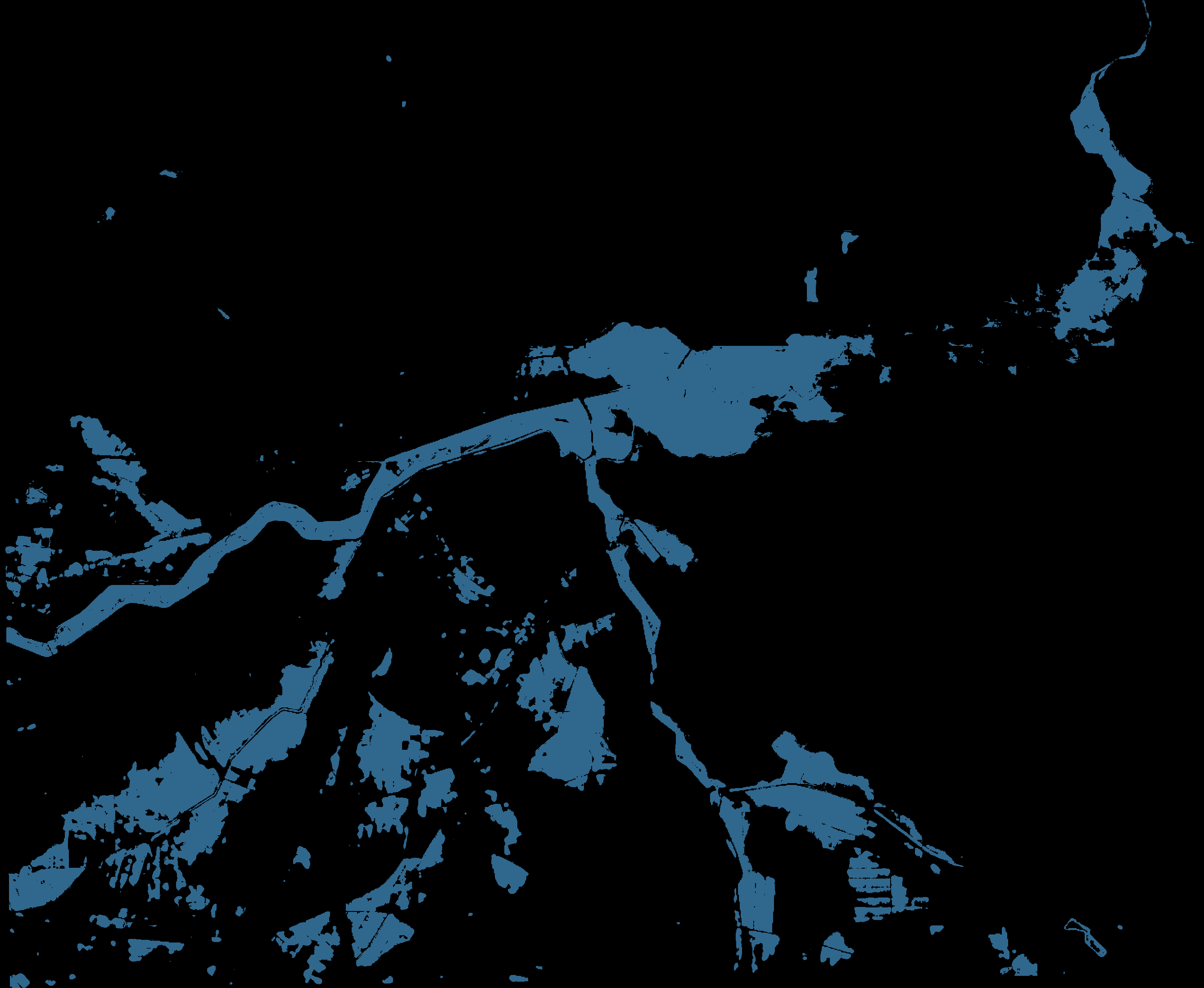}}\hfil

    \medskip
    \subfloat[]{\includegraphics[width=0.15\textwidth]{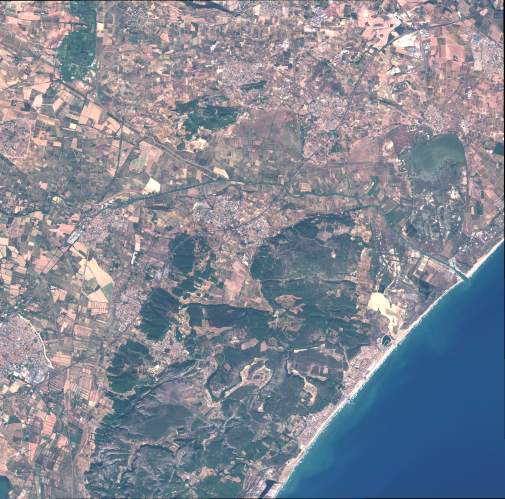}}\hfil
    \subfloat[]{\includegraphics[width=0.15\textwidth]{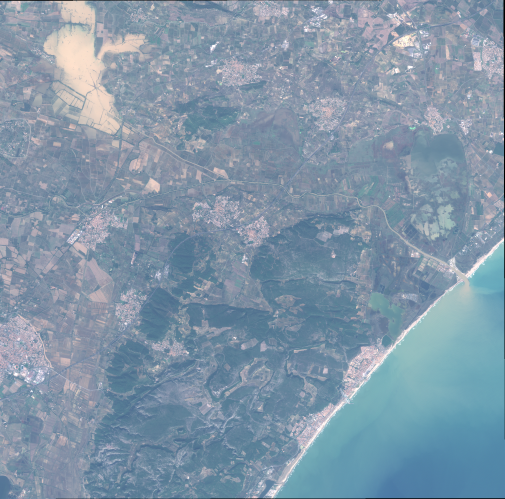}}\hfil
    \subfloat[]{\includegraphics[width=0.15\textwidth]{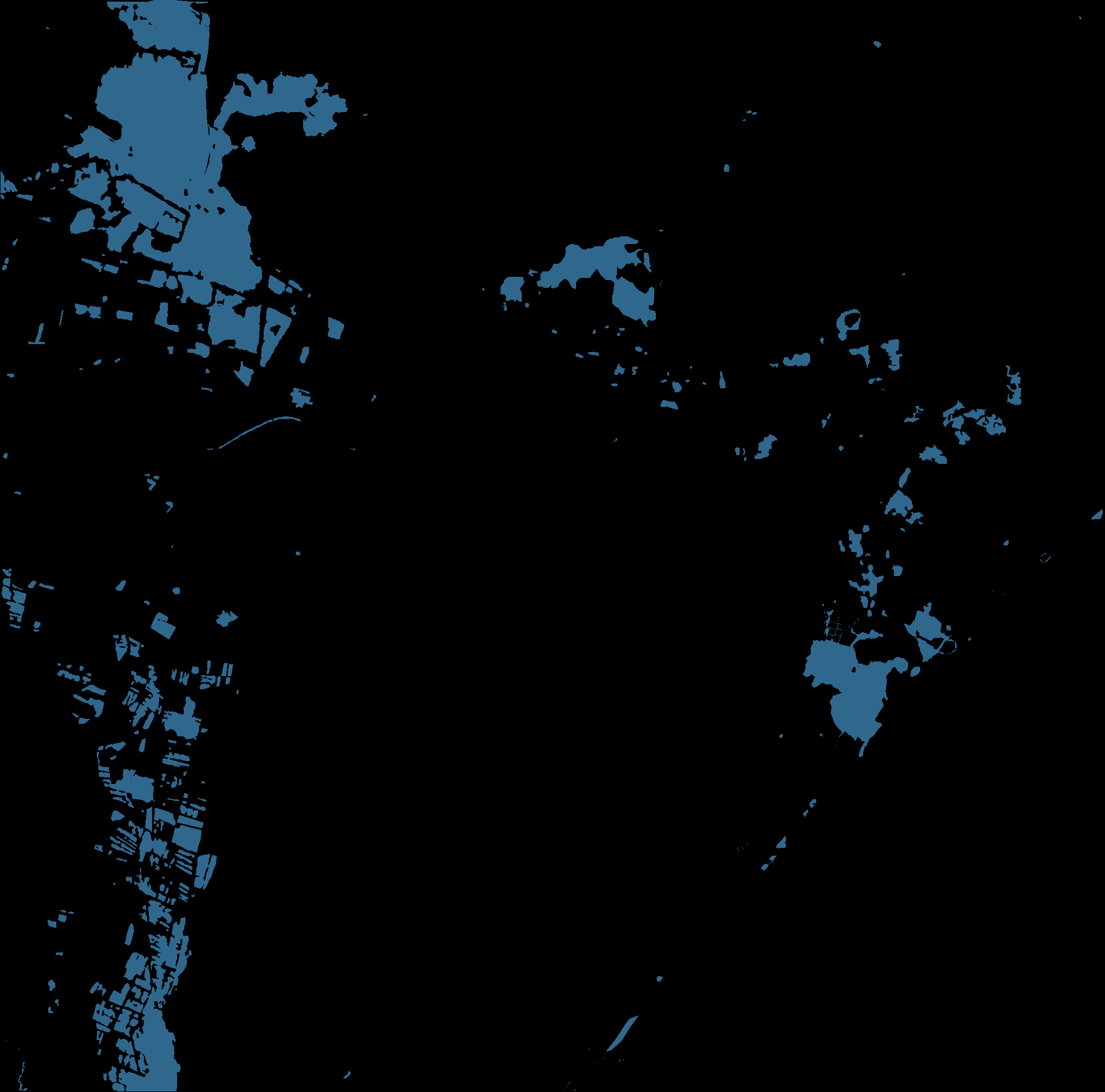}}\hfil
    \subfloat[]{\includegraphics[width=0.15\textwidth]{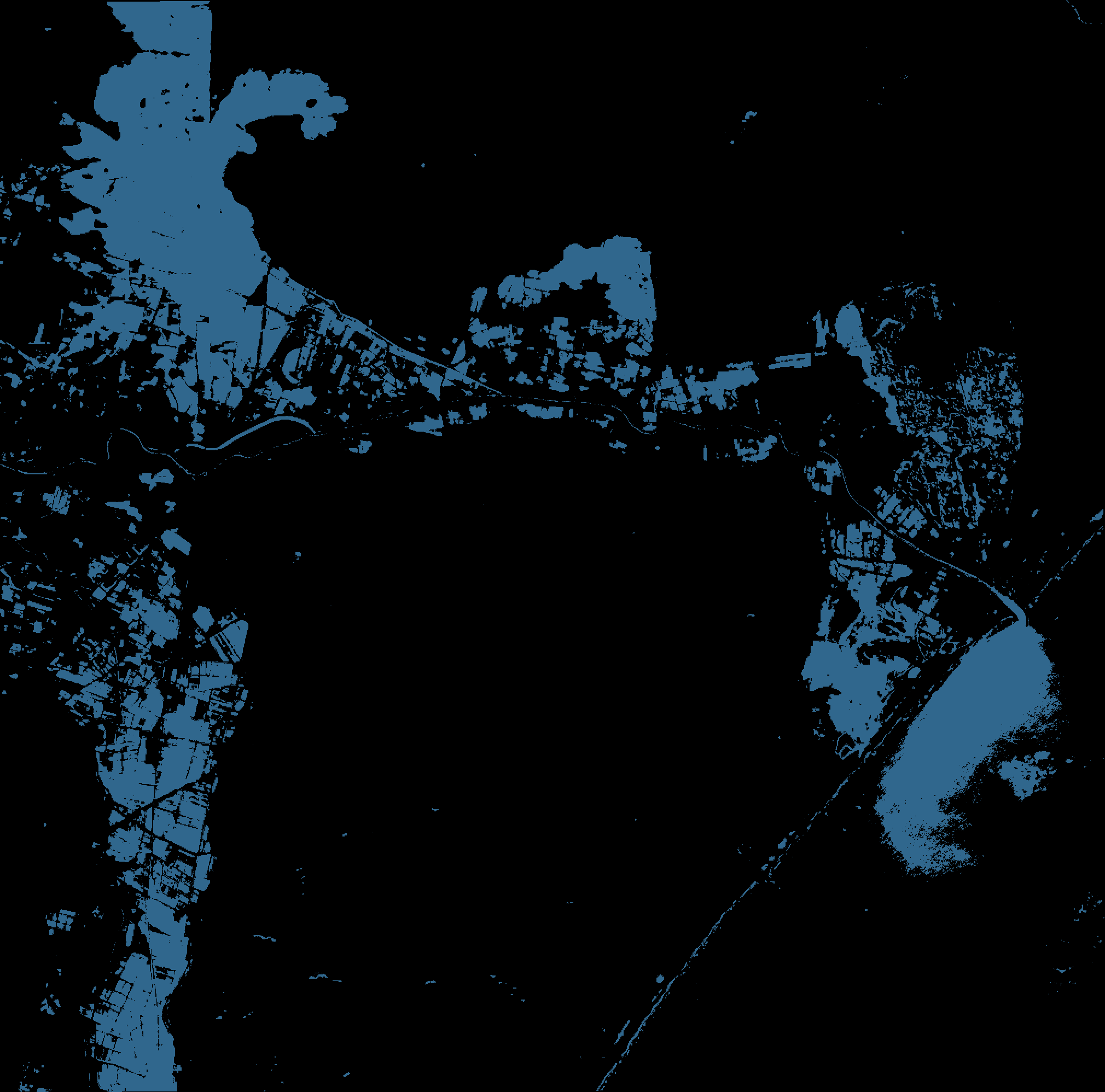}}\hfil
    \subfloat[]{\includegraphics[width=0.15\textwidth]{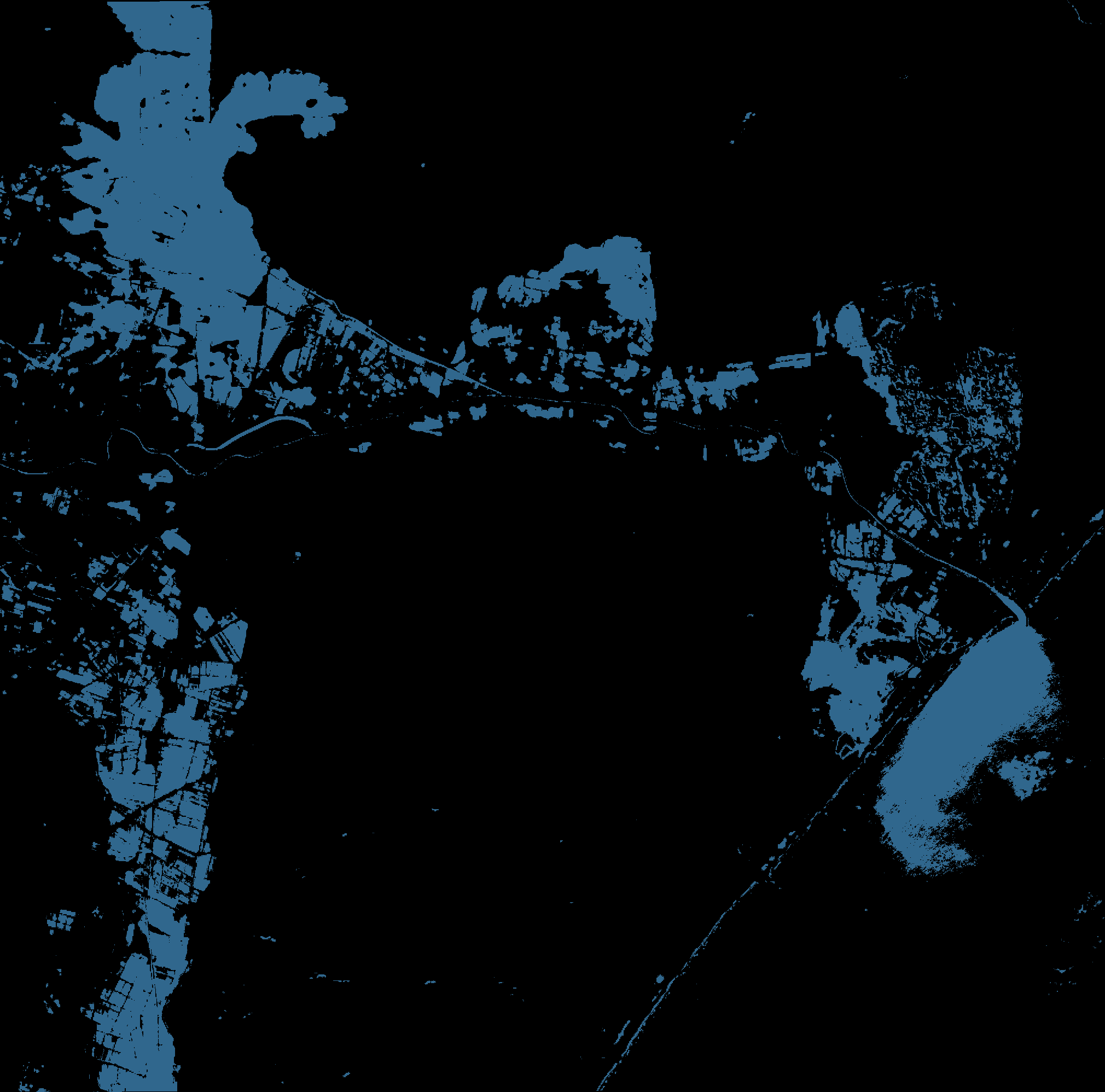}}\hfil

    \medskip
    \subfloat[Pre Flood]{\includegraphics[width=0.15\textwidth]{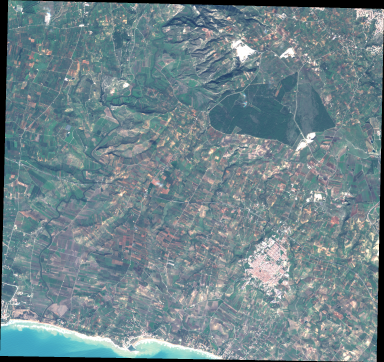}}\hfil
    \subfloat[After Flood]{\includegraphics[width=0.15\textwidth]{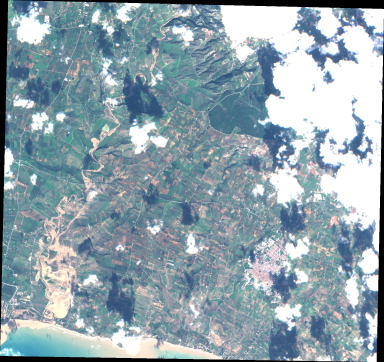}}\hfil
    \subfloat[New Water (ground truth]{\includegraphics[width=0.15\textwidth]{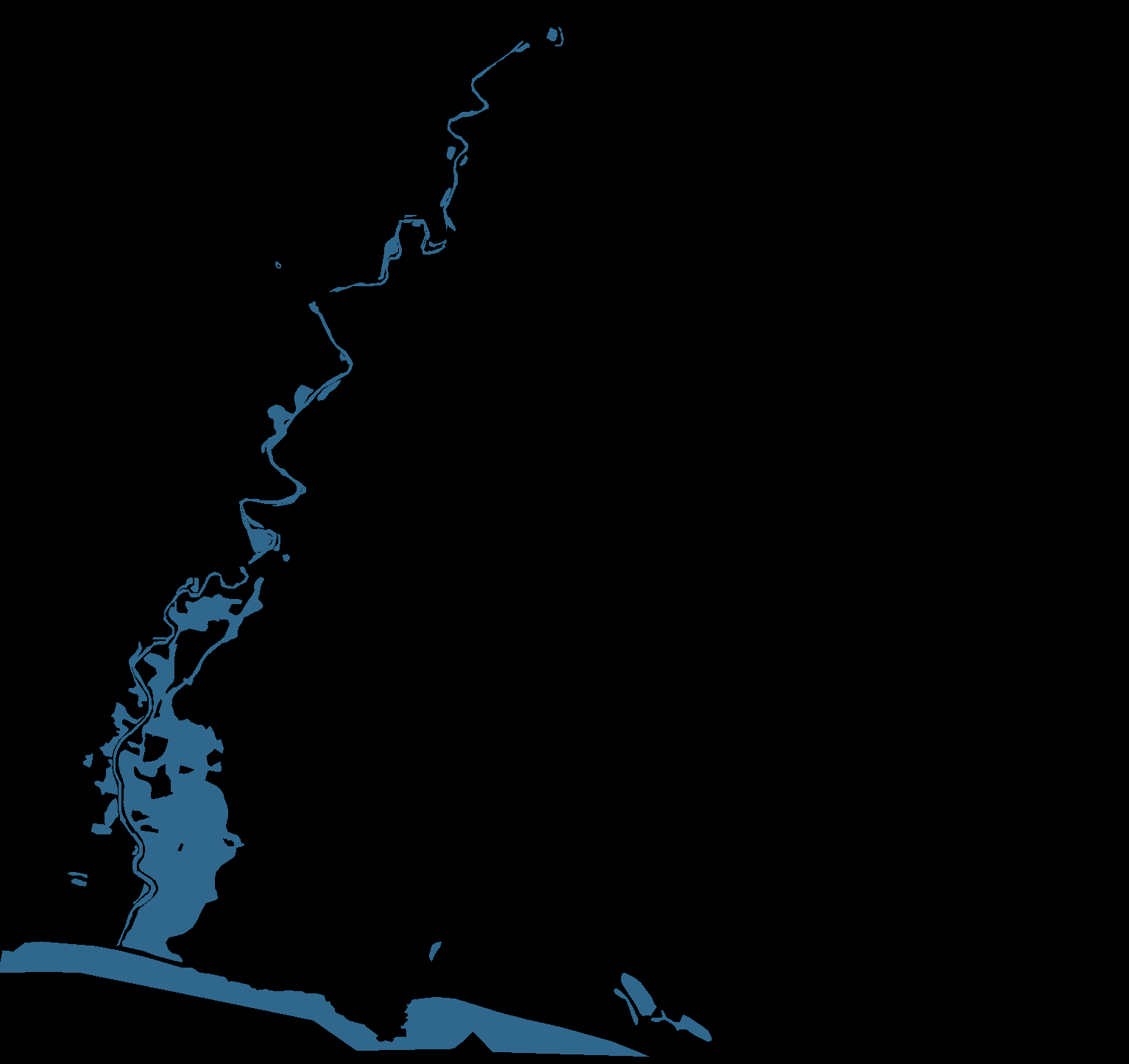}}\hfil
    \subfloat[U-Net]{\includegraphics[width=0.15\textwidth]{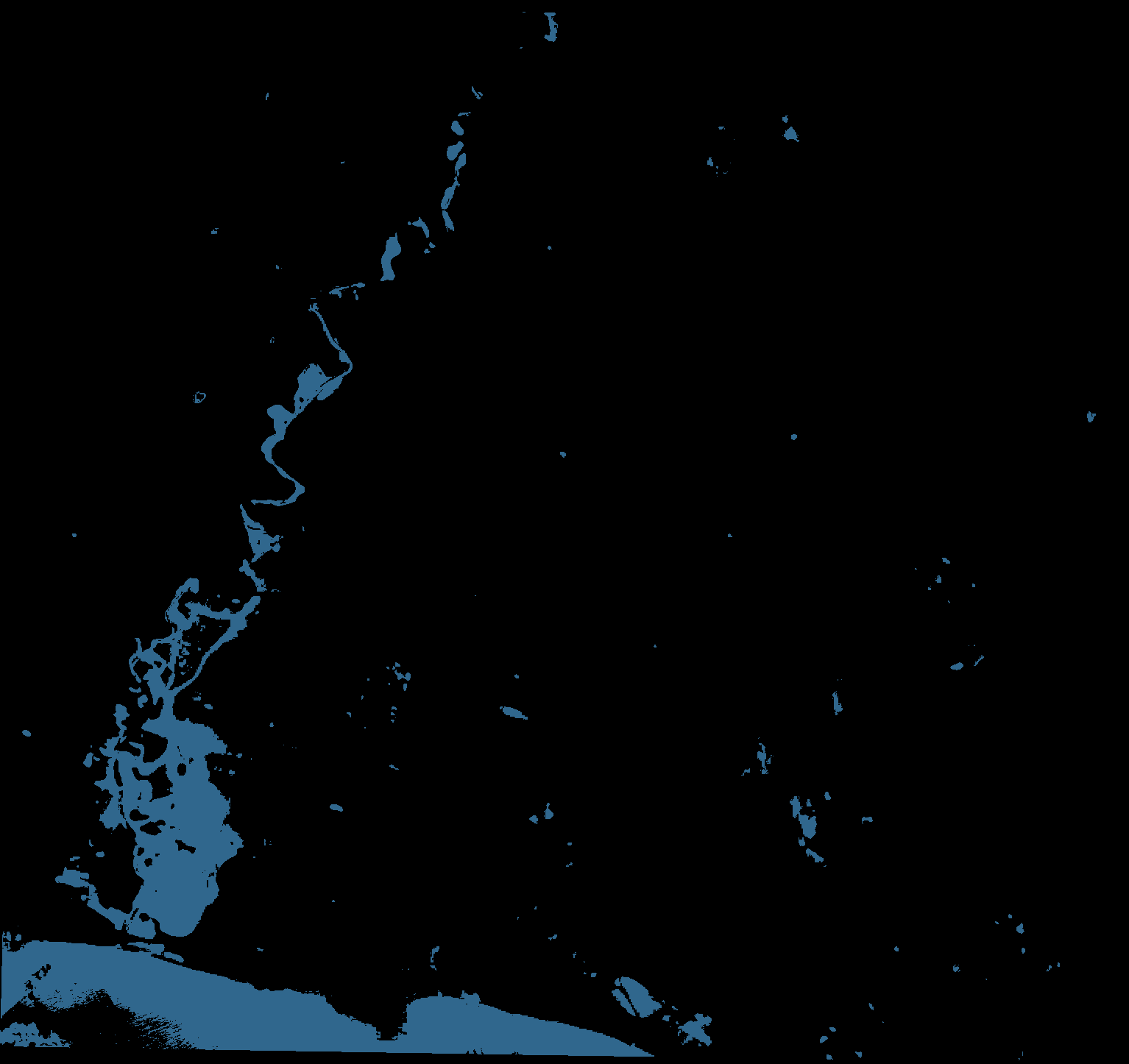}}\hfil
    \subfloat[IDSS+]
    {\includegraphics[width=0.15\textwidth]{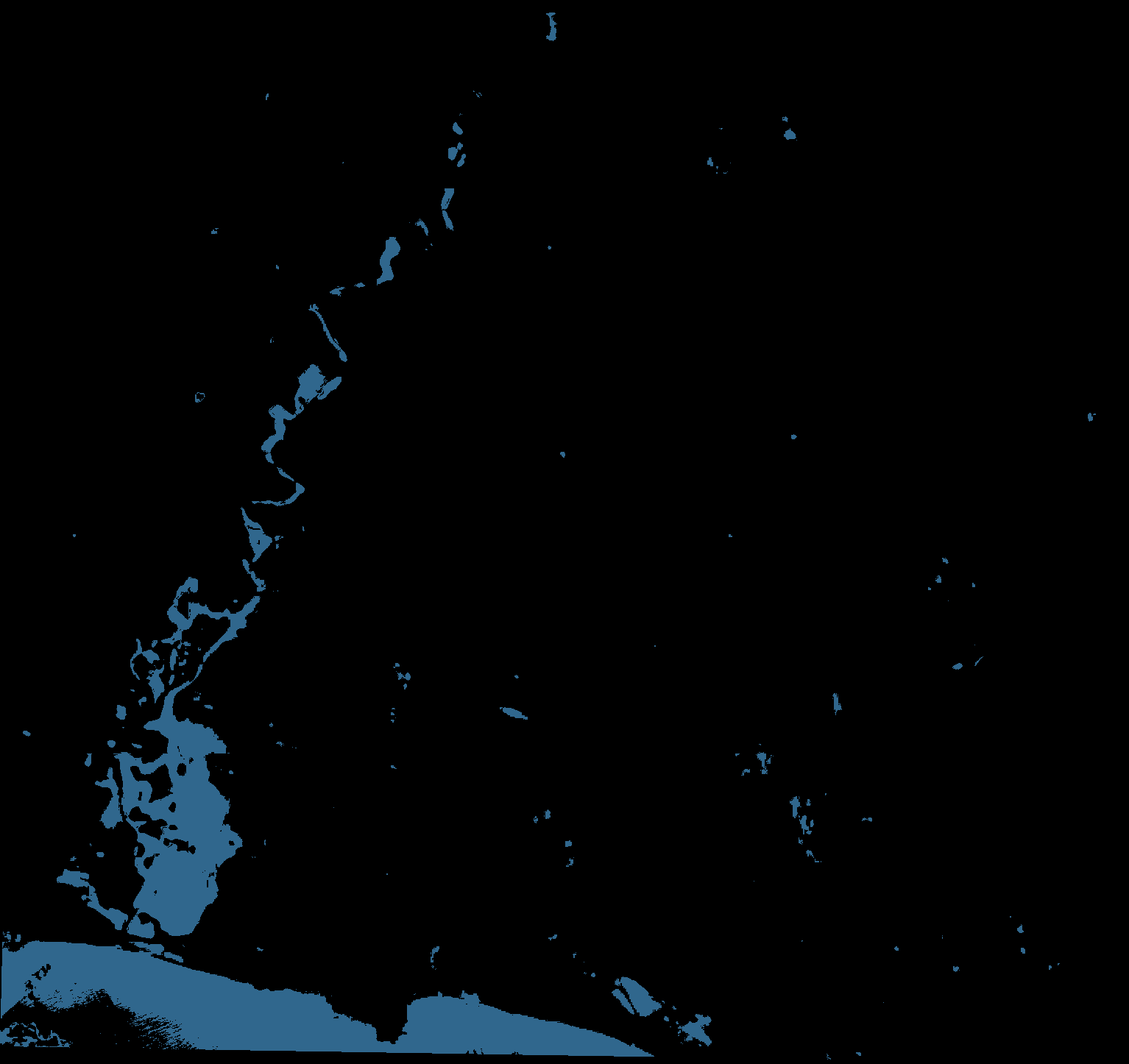}}

    \caption{Comparison of segmentation change detection results. The blue color in Ground Truth, IMAFD (U-Net) and IMAFD (IDSS+) indicate changed water/flood.}
    \label{semantic_change_results}
\end{figure*}

2) Semantic segmentation

To evaluate the effectiveness of the IDSS+ semantic segmentation, we conduct experiments on the WorldFloods dataset and compare them with other methods, as shown in Table \ref{tab:IoU comparison}. It can be observed that the proposed IDSS+ provides the highest IoU water and mIoU, outperforming the benchmark U-Net and other methods. Specifically, the IoU water is exceeded by $1.57\%$ and the mIoU is exceeded by $0.19\%$ compared to U-Net. In particular, it can be seen from Figure \ref{noise} that the prediction map of IDSS+ has less noise compared to U-Net.

Furthermore, the prototype in the latent space was visualized with the number of prototypes for each class equal to 10 and 100, as shown in Figure \ref{UMAP}. It can be observed that 100 prototypes for each class have a better separation compared to 10 prototypes for each class. This also explains the better performance of the former as shown in the Table \ref{tab:K ablation study}.

In addition, to better visualize the decision-making process of the algorithm, the UMAP plot could be drawn based on the visulization of the prototypes in the latent features to detail the prediction procedure of each pixel and the reason for the misclassification. As shown in figure \ref{correct_pre}, the prototypes in the latent feature space are marked as three different coloured circles. It can be seen that the unknown red triangular pixel is successfully predicted as Water as it is surrounded by blue circles representing Water prototypes. Comparatively, in figure \ref{incorrect_pre}, the unknown red triangular pixel is predicted as Land as it is surrounded by green Land prototypes, even though its ground truth label is  Water. This explains why the unknown pixel was incorrectly predicted as Land.

3) Semantic change detection

The ablation study of the semantic change detection was assessed on the RavAEn dataset. The results, shown in Table \ref{semantic_change_results_table} compare the performance of IDSS+  and U-Net when labeling the binary change maps obtained from the second stage of IMAFD. It can be observed that IDSS+ achieves competitive results compared to U-Net. Specifically, in most scenarios, IDSS+ surpasses U-Net in terms of Precision, F1, and IoU water, while U-Net achieves better results in terms of Recall. This shows that the U-Net is more prone to overpredicting flood extent. The visualized results are shown in Figure \ref{semantic_change_results}.

\section{Conclusion}

In this paper, we proposed a holistic, efficient, and interpretable flood detection framework, IMAFD. Our approach divides flood detection into four steps and progressively narrows down the flood detection problem, starting with detecting suspected images from the time series moving to detect the semantic change at the pixel level. In addition, the semantic change detection (stage 3 of IMAFD) method IDSS+ provides additional insight based on UMAP plots and confidence maps to help the users further understand the decision-making process of the algorithm. Finally, the comparative experiments and ablation study were performed on WorldFloods RavAEn and MediaEvfal datasets, demonstrating the effectiveness and efficiency of the proposed method.

\section*{Reproducibility}

In this work, the U-Net model was trained on the Wordflooods dataset and can be accessed from the official GitHub repository of the paper \citep{mateo2021towards}. The DINO-ViT-S/16 model was pre-trained on the SSL4EO-S12 dataset \citep{ wang2022ssl4eo} and can be accessed from Torchgeo. The k value in mini-batch k-means and k nearest neighbour have been set at 100 and 10, respectively, after multiple tests to give the best results.

\section*{Acknowledgements}

This work has been funded by the European Space Agency (contract 4000133854/21/NL/CBi) and supported by the ESA $\phi$ – lab. This work is also supported by ELSA – European Lighthouse on Secure and Safe AI funded by the European Union under grant agreement No. 101070617. Views and opinions expressed are however those of the author(s) only and do not necessarily reflect those of the European Union or European Commission. Neither the European Union nor the European Commission can be held responsible.

The computational experiments were supported by a High-End Computing (HEC) Cluster of Lancaster University.

\bibliographystyle{elsarticle-harv} 
\bibliography{bio}

\begin{thebibliography}{61}
\expandafter\ifx\csname natexlab\endcsname\relax\def\natexlab#1{#1}\fi
\providecommand{\url}[1]{\texttt{#1}}
\providecommand{\href}[2]{#2}
\providecommand{\path}[1]{#1}
\providecommand{\DOIprefix}{doi:}
\providecommand{\ArXivprefix}{arXiv:}
\providecommand{\URLprefix}{URL: }
\providecommand{\Pubmedprefix}{pmid:}
\providecommand{\doi}[1]{\href{http://dx.doi.org/#1}{\path{#1}}}
\providecommand{\Pubmed}[1]{\href{pmid:#1}{\path{#1}}}
\providecommand{\bibinfo}[2]{#2}
\ifx\xfnm\relax \def\xfnm[#1]{\unskip,\space#1}\fi
%Type = Article
\bibitem[{Al-Amri et~al.(2010)Al-Amri, Kalyankar et~al.}]{al2010image}
\bibinfo{author}{Al-Amri, S.S.}, \bibinfo{author}{Kalyankar, N.V.}, et~al., \bibinfo{year}{2010}.
\newblock \bibinfo{title}{Image segmentation by using threshold techniques}.
\newblock \bibinfo{journal}{arXiv preprint arXiv:1005.4020} .
%Type = Book
\bibitem[{Angelov(2012)}]{angelov2012autonomous}
\bibinfo{author}{Angelov, P.}, \bibinfo{year}{2012}.
\newblock \bibinfo{title}{Autonomous learning systems: from data streams to knowledge in real-time}.
\newblock \bibinfo{publisher}{John Wiley \& Sons}.
%Type = Article
\bibitem[{Angelov et~al.(2023)Angelov, Kangin and Zhang}]{angelov2023towards}
\bibinfo{author}{Angelov, P.}, \bibinfo{author}{Kangin, D.}, \bibinfo{author}{Zhang, Z.}, \bibinfo{year}{2023}.
\newblock \bibinfo{title}{Towards interpretable-by-design deep learning algorithms}.
\newblock \bibinfo{journal}{arXiv preprint arXiv:2311.11396} .
%Type = Article
\bibitem[{Angelov and Soares(2020)}]{angelov2020towards}
\bibinfo{author}{Angelov, P.}, \bibinfo{author}{Soares, E.}, \bibinfo{year}{2020}.
\newblock \bibinfo{title}{Towards explainable deep neural networks (xdnn)}.
\newblock \bibinfo{journal}{Neural Networks} \bibinfo{volume}{130}, \bibinfo{pages}{185--194}.
%Type = Article
\bibitem[{Badrinarayanan et~al.(2017)Badrinarayanan, Kendall and Cipolla}]{badrinarayanan2017segnet}
\bibinfo{author}{Badrinarayanan, V.}, \bibinfo{author}{Kendall, A.}, \bibinfo{author}{Cipolla, R.}, \bibinfo{year}{2017}.
\newblock \bibinfo{title}{Segnet: A deep convolutional encoder-decoder architecture for image segmentation}.
\newblock \bibinfo{journal}{IEEE transactions on pattern analysis and machine intelligence} \bibinfo{volume}{39}, \bibinfo{pages}{2481--2495}.
%Type = Article
\bibitem[{Bai et~al.(2023)Bai, Wang, Yin, Sun, Chen, Li and Li}]{bai2023deep}
\bibinfo{author}{Bai, T.}, \bibinfo{author}{Wang, L.}, \bibinfo{author}{Yin, D.}, \bibinfo{author}{Sun, K.}, \bibinfo{author}{Chen, Y.}, \bibinfo{author}{Li, W.}, \bibinfo{author}{Li, D.}, \bibinfo{year}{2023}.
\newblock \bibinfo{title}{Deep learning for change detection in remote sensing: a review}.
\newblock \bibinfo{journal}{Geo-spatial Information Science} \bibinfo{volume}{26}, \bibinfo{pages}{262--288}.
%Type = Article
\bibitem[{Biehl et~al.(2016)Biehl, Hammer and Villmann}]{biehl2016prototype}
\bibinfo{author}{Biehl, M.}, \bibinfo{author}{Hammer, B.}, \bibinfo{author}{Villmann, T.}, \bibinfo{year}{2016}.
\newblock \bibinfo{title}{Prototype-based models in machine learning}.
\newblock \bibinfo{journal}{Wiley Interdisciplinary Reviews: Cognitive Science} \bibinfo{volume}{7}, \bibinfo{pages}{92--111}.
%Type = Inproceedings
\bibitem[{Bischke et~al.(2019)Bischke, Helber, Brugman, Basar, Zhao, Larson and Pogorelov}]{bischke2019multimedia}
\bibinfo{author}{Bischke, B.}, \bibinfo{author}{Helber, P.}, \bibinfo{author}{Brugman, S.}, \bibinfo{author}{Basar, E.}, \bibinfo{author}{Zhao, Z.}, \bibinfo{author}{Larson, M.A.}, \bibinfo{author}{Pogorelov, K.}, \bibinfo{year}{2019}.
\newblock \bibinfo{title}{The multimedia satellite task at mediaeval 2019.}, in: \bibinfo{booktitle}{MediaEval}.
%Type = Inproceedings
\bibitem[{B{\"o}hle et~al.(2022)B{\"o}hle, Fritz and Schiele}]{bohle2022b}
\bibinfo{author}{B{\"o}hle, M.}, \bibinfo{author}{Fritz, M.}, \bibinfo{author}{Schiele, B.}, \bibinfo{year}{2022}.
\newblock \bibinfo{title}{B-cos networks: Alignment is all we need for interpretability}, in: \bibinfo{booktitle}{Proceedings of the IEEE/CVF Conference on Computer Vision and Pattern Recognition}, pp. \bibinfo{pages}{10329--10338}.
%Type = Article
\bibitem[{B{\"o}hle et~al.(2024)B{\"o}hle, Singh, Fritz and Schiele}]{bohle2024b}
\bibinfo{author}{B{\"o}hle, M.}, \bibinfo{author}{Singh, N.}, \bibinfo{author}{Fritz, M.}, \bibinfo{author}{Schiele, B.}, \bibinfo{year}{2024}.
\newblock \bibinfo{title}{B-cos alignment for inherently interpretable cnns and vision transformers}.
\newblock \bibinfo{journal}{IEEE Transactions on Pattern Analysis and Machine Intelligence} .
%Type = Article
\bibitem[{Chan et~al.(2022)Chan, Kong and Liang}]{chan2022comparative}
\bibinfo{author}{Chan, C.S.}, \bibinfo{author}{Kong, H.}, \bibinfo{author}{Liang, G.}, \bibinfo{year}{2022}.
\newblock \bibinfo{title}{A comparative study of faithfulness metrics for model interpretability methods}.
\newblock \bibinfo{journal}{arXiv preprint arXiv:2204.05514} .
%Type = Article
\bibitem[{Chen et~al.(2019)Chen, Li, Tao, Barnett, Rudin and Su}]{chen2019looks}
\bibinfo{author}{Chen, C.}, \bibinfo{author}{Li, O.}, \bibinfo{author}{Tao, D.}, \bibinfo{author}{Barnett, A.}, \bibinfo{author}{Rudin, C.}, \bibinfo{author}{Su, J.K.}, \bibinfo{year}{2019}.
\newblock \bibinfo{title}{This looks like that: deep learning for interpretable image recognition}.
\newblock \bibinfo{journal}{Advances in neural information processing systems} \bibinfo{volume}{32}.
%Type = Article
\bibitem[{Chen et~al.(2021)Chen, Qi and Shi}]{chen2021remote}
\bibinfo{author}{Chen, H.}, \bibinfo{author}{Qi, Z.}, \bibinfo{author}{Shi, Z.}, \bibinfo{year}{2021}.
\newblock \bibinfo{title}{Remote sensing image change detection with transformers}.
\newblock \bibinfo{journal}{IEEE Transactions on Geoscience and Remote Sensing} \bibinfo{volume}{60}, \bibinfo{pages}{1--14}.
%Type = Inproceedings
\bibitem[{Chen et~al.(2018)Chen, Zhu, Papandreou, Schroff and Adam}]{chen2018encoder}
\bibinfo{author}{Chen, L.C.}, \bibinfo{author}{Zhu, Y.}, \bibinfo{author}{Papandreou, G.}, \bibinfo{author}{Schroff, F.}, \bibinfo{author}{Adam, H.}, \bibinfo{year}{2018}.
\newblock \bibinfo{title}{Encoder-decoder with atrous separable convolution for semantic image segmentation}, in: \bibinfo{booktitle}{Proceedings of the European conference on computer vision (ECCV)}, pp. \bibinfo{pages}{801--818}.
%Type = Article
\bibitem[{Clare et~al.(2022)Clare, Sonnewald, Lguensat, Deshayes and Balaji}]{clare2022explainable}
\bibinfo{author}{Clare, M.C.}, \bibinfo{author}{Sonnewald, M.}, \bibinfo{author}{Lguensat, R.}, \bibinfo{author}{Deshayes, J.}, \bibinfo{author}{Balaji, V.}, \bibinfo{year}{2022}.
\newblock \bibinfo{title}{Explainable artificial intelligence for bayesian neural networks: toward trustworthy predictions of ocean dynamics}.
\newblock \bibinfo{journal}{Journal of Advances in Modeling Earth Systems} \bibinfo{volume}{14}, \bibinfo{pages}{e2022MS003162}.
%Type = Article
\bibitem[{Dosovitskiy et~al.(2020)Dosovitskiy, Beyer, Kolesnikov, Weissenborn, Zhai, Unterthiner, Dehghani, Minderer, Heigold, Gelly et~al.}]{dosovitskiy2020image}
\bibinfo{author}{Dosovitskiy, A.}, \bibinfo{author}{Beyer, L.}, \bibinfo{author}{Kolesnikov, A.}, \bibinfo{author}{Weissenborn, D.}, \bibinfo{author}{Zhai, X.}, \bibinfo{author}{Unterthiner, T.}, \bibinfo{author}{Dehghani, M.}, \bibinfo{author}{Minderer, M.}, \bibinfo{author}{Heigold, G.}, \bibinfo{author}{Gelly, S.}, et~al., \bibinfo{year}{2020}.
\newblock \bibinfo{title}{An image is worth 16x16 words: Transformers for image recognition at scale}.
\newblock \bibinfo{journal}{arXiv preprint arXiv:2010.11929} .
%Type = Article
\bibitem[{Gao(1996)}]{gao1996ndwi}
\bibinfo{author}{Gao, B.C.}, \bibinfo{year}{1996}.
\newblock \bibinfo{title}{Ndwi—a normalized difference water index for remote sensing of vegetation liquid water from space}.
\newblock \bibinfo{journal}{Remote sensing of environment} \bibinfo{volume}{58}, \bibinfo{pages}{257--266}.
%Type = Article
\bibitem[{Gevaert(2022)}]{gevaert2022explainable}
\bibinfo{author}{Gevaert, C.M.}, \bibinfo{year}{2022}.
\newblock \bibinfo{title}{Explainable ai for earth observation: A review including societal and regulatory perspectives}.
\newblock \bibinfo{journal}{International Journal of Applied Earth Observation and Geoinformation} \bibinfo{volume}{112}, \bibinfo{pages}{102869}.
%Type = Article
\bibitem[{Gong et~al.(2011)Gong, Cao and Wu}]{gong2011neighborhood}
\bibinfo{author}{Gong, M.}, \bibinfo{author}{Cao, Y.}, \bibinfo{author}{Wu, Q.}, \bibinfo{year}{2011}.
\newblock \bibinfo{title}{A neighborhood-based ratio approach for change detection in sar images}.
\newblock \bibinfo{journal}{IEEE Geoscience and Remote Sensing Letters} \bibinfo{volume}{9}, \bibinfo{pages}{307--311}.
%Type = Article
\bibitem[{Guo et~al.(2022)Guo, Hou, Wu, Ren, Wang and Jiao}]{guo2022prob}
\bibinfo{author}{Guo, X.}, \bibinfo{author}{Hou, B.}, \bibinfo{author}{Wu, Z.}, \bibinfo{author}{Ren, B.}, \bibinfo{author}{Wang, S.}, \bibinfo{author}{Jiao, L.}, \bibinfo{year}{2022}.
\newblock \bibinfo{title}{Prob-pos: A framework for improving visual explanations from convolutional neural networks for remote sensing image classification}.
\newblock \bibinfo{journal}{Remote Sensing} \bibinfo{volume}{14}, \bibinfo{pages}{3042}.
%Type = Article
\bibitem[{Hao et~al.(2023)Hao, Yang, Lin, Zou, Liu and Zhang}]{hao2023bi}
\bibinfo{author}{Hao, M.}, \bibinfo{author}{Yang, C.}, \bibinfo{author}{Lin, H.}, \bibinfo{author}{Zou, L.}, \bibinfo{author}{Liu, S.}, \bibinfo{author}{Zhang, H.}, \bibinfo{year}{2023}.
\newblock \bibinfo{title}{Bi-temporal change detection of high-resolution images by referencing time series medium-resolution images}.
\newblock \bibinfo{journal}{International Journal of Remote Sensing} \bibinfo{volume}{44}, \bibinfo{pages}{3333--3357}.
%Type = Article
\bibitem[{Hung et~al.(2020)Hung, Wu and Tseng}]{hung2020remote}
\bibinfo{author}{Hung, S.C.}, \bibinfo{author}{Wu, H.C.}, \bibinfo{author}{Tseng, M.H.}, \bibinfo{year}{2020}.
\newblock \bibinfo{title}{Remote sensing scene classification and explanation using rsscnet and lime}.
\newblock \bibinfo{journal}{Applied Sciences} \bibinfo{volume}{10}, \bibinfo{pages}{6151}.
%Type = Article
\bibitem[{Iqbal et~al.(2021)Iqbal, Perez, Li and Barthelemy}]{iqbal2021computer}
\bibinfo{author}{Iqbal, U.}, \bibinfo{author}{Perez, P.}, \bibinfo{author}{Li, W.}, \bibinfo{author}{Barthelemy, J.}, \bibinfo{year}{2021}.
\newblock \bibinfo{title}{How computer vision can facilitate flood management: A systematic review}.
\newblock \bibinfo{journal}{International Journal of Disaster Risk Reduction} \bibinfo{volume}{53}, \bibinfo{pages}{102030}.
%Type = Article
\bibitem[{Jongman et~al.(2012)Jongman, Ward and Aerts}]{jongman2012global}
\bibinfo{author}{Jongman, B.}, \bibinfo{author}{Ward, P.J.}, \bibinfo{author}{Aerts, J.C.}, \bibinfo{year}{2012}.
\newblock \bibinfo{title}{Global exposure to river and coastal flooding: Long term trends and changes}.
\newblock \bibinfo{journal}{Global Environmental Change} \bibinfo{volume}{22}, \bibinfo{pages}{823--835}.
%Type = Article
\bibitem[{Katiyar et~al.(2021)Katiyar, Tamkuan and Nagai}]{katiyar2021near}
\bibinfo{author}{Katiyar, V.}, \bibinfo{author}{Tamkuan, N.}, \bibinfo{author}{Nagai, M.}, \bibinfo{year}{2021}.
\newblock \bibinfo{title}{Near-real-time flood mapping using off-the-shelf models with sar imagery and deep learning}.
\newblock \bibinfo{journal}{Remote Sensing} \bibinfo{volume}{13}, \bibinfo{pages}{2334}.
%Type = Article
\bibitem[{Kaufman(1990)}]{kaufman1990partitioning}
\bibinfo{author}{Kaufman, L.}, \bibinfo{year}{1990}.
\newblock \bibinfo{title}{Partitioning around medoids (program pam)}.
\newblock \bibinfo{journal}{Finding groups in data} \bibinfo{volume}{344}, \bibinfo{pages}{68--125}.
%Type = Article
\bibitem[{Khelifi and Mignotte(2020)}]{khelifi2020deep}
\bibinfo{author}{Khelifi, L.}, \bibinfo{author}{Mignotte, M.}, \bibinfo{year}{2020}.
\newblock \bibinfo{title}{Deep learning for change detection in remote sensing images: Comprehensive review and meta-analysis}.
\newblock \bibinfo{journal}{Ieee Access} \bibinfo{volume}{8}, \bibinfo{pages}{126385--126400}.
%Type = Article
\bibitem[{Landuyt et~al.(2020)Landuyt, Verhoest and Van~Coillie}]{landuyt2020flood}
\bibinfo{author}{Landuyt, L.}, \bibinfo{author}{Verhoest, N.E.}, \bibinfo{author}{Van~Coillie, F.M.}, \bibinfo{year}{2020}.
\newblock \bibinfo{title}{Flood mapping in vegetated areas using an unsupervised clustering approach on sentinel-1 and-2 imagery}.
\newblock \bibinfo{journal}{Remote Sensing} \bibinfo{volume}{12}, \bibinfo{pages}{3611}.
%Type = Article
\bibitem[{Liu et~al.(2015)Liu, Bruzzone, Bovolo, Zanetti and Du}]{liu2015sequential}
\bibinfo{author}{Liu, S.}, \bibinfo{author}{Bruzzone, L.}, \bibinfo{author}{Bovolo, F.}, \bibinfo{author}{Zanetti, M.}, \bibinfo{author}{Du, P.}, \bibinfo{year}{2015}.
\newblock \bibinfo{title}{Sequential spectral change vector analysis for iteratively discovering and detecting multiple changes in hyperspectral images}.
\newblock \bibinfo{journal}{IEEE transactions on geoscience and remote sensing} \bibinfo{volume}{53}, \bibinfo{pages}{4363--4378}.
%Type = Inproceedings
\bibitem[{Long et~al.(2015)Long, Shelhamer and Darrell}]{long2015fully}
\bibinfo{author}{Long, J.}, \bibinfo{author}{Shelhamer, E.}, \bibinfo{author}{Darrell, T.}, \bibinfo{year}{2015}.
\newblock \bibinfo{title}{Fully convolutional networks for semantic segmentation}, in: \bibinfo{booktitle}{Proceedings of the IEEE conference on computer vision and pattern recognition}, pp. \bibinfo{pages}{3431--3440}.
%Type = Article
\bibitem[{Long et~al.(2014)Long, Fatoyinbo and Policelli}]{long2014flood}
\bibinfo{author}{Long, S.}, \bibinfo{author}{Fatoyinbo, T.E.}, \bibinfo{author}{Policelli, F.}, \bibinfo{year}{2014}.
\newblock \bibinfo{title}{Flood extent mapping for namibia using change detection and thresholding with sar}.
\newblock \bibinfo{journal}{Environmental Research Letters} \bibinfo{volume}{9}, \bibinfo{pages}{035002}.
%Type = Inproceedings
\bibitem[{MacQueen et~al.(1967)}]{macqueen1967some}
\bibinfo{author}{MacQueen, J.}, et~al., \bibinfo{year}{1967}.
\newblock \bibinfo{title}{Some methods for classification and analysis of multivariate observations}, in: \bibinfo{booktitle}{Proceedings of the fifth Berkeley symposium on mathematical statistics and probability}, \bibinfo{organization}{Oakland, CA, USA}. pp. \bibinfo{pages}{281--297}.
%Type = Inproceedings
\bibitem[{Malila(1980)}]{malila1980change}
\bibinfo{author}{Malila, W.A.}, \bibinfo{year}{1980}.
\newblock \bibinfo{title}{Change vector analysis: An approach for detecting forest changes with landsat}, in: \bibinfo{booktitle}{LARS symposia}, p. \bibinfo{pages}{385}.
%Type = Article
\bibitem[{Mall et~al.(2022)Mall, Hariharan and Bala}]{mall2022change}
\bibinfo{author}{Mall, U.}, \bibinfo{author}{Hariharan, B.}, \bibinfo{author}{Bala, K.}, \bibinfo{year}{2022}.
\newblock \bibinfo{title}{Change event dataset for discovery from spatio-temporal remote sensing imagery}.
\newblock \bibinfo{journal}{Advances in Neural Information Processing Systems} \bibinfo{volume}{35}, \bibinfo{pages}{27484--27496}.
%Type = Article
\bibitem[{Mateo-Garcia et~al.(2021)Mateo-Garcia, Veitch-Michaelis, Smith, Oprea, Schumann, Gal, Baydin and Backes}]{mateo2021towards}
\bibinfo{author}{Mateo-Garcia, G.}, \bibinfo{author}{Veitch-Michaelis, J.}, \bibinfo{author}{Smith, L.}, \bibinfo{author}{Oprea, S.V.}, \bibinfo{author}{Schumann, G.}, \bibinfo{author}{Gal, Y.}, \bibinfo{author}{Baydin, A.G.}, \bibinfo{author}{Backes, D.}, \bibinfo{year}{2021}.
\newblock \bibinfo{title}{Towards global flood mapping onboard low cost satellites with machine learning}.
\newblock \bibinfo{journal}{Scientific reports} \bibinfo{volume}{11}, \bibinfo{pages}{1--12}.
%Type = Article
\bibitem[{Mitros and Mac~Namee(2019)}]{mitros2019validity}
\bibinfo{author}{Mitros, J.}, \bibinfo{author}{Mac~Namee, B.}, \bibinfo{year}{2019}.
\newblock \bibinfo{title}{On the validity of bayesian neural networks for uncertainty estimation}.
\newblock \bibinfo{journal}{arXiv preprint arXiv:1912.01530} .
%Type = Article
\bibitem[{Pan et~al.(2020)Pan, Xi and Wang}]{pan2020comparative}
\bibinfo{author}{Pan, F.}, \bibinfo{author}{Xi, X.}, \bibinfo{author}{Wang, C.}, \bibinfo{year}{2020}.
\newblock \bibinfo{title}{A comparative study of water indices and image classification algorithms for mapping inland surface water bodies using landsat imagery}.
\newblock \bibinfo{journal}{Remote Sensing} \bibinfo{volume}{12}, \bibinfo{pages}{1611}.
%Type = Article
\bibitem[{Regulation(2016)}]{regulation2016regulation}
\bibinfo{author}{Regulation, P.}, \bibinfo{year}{2016}.
\newblock \bibinfo{title}{Regulation (eu) 2016/679 of the european parliament and of the council}.
\newblock \bibinfo{journal}{Regulation (eu)} \bibinfo{volume}{679}, \bibinfo{pages}{2016}.
%Type = Inproceedings
\bibitem[{Ribeiro et~al.(2016)Ribeiro, Singh and Guestrin}]{ribeiro2016should}
\bibinfo{author}{Ribeiro, M.T.}, \bibinfo{author}{Singh, S.}, \bibinfo{author}{Guestrin, C.}, \bibinfo{year}{2016}.
\newblock \bibinfo{title}{" why should i trust you?" explaining the predictions of any classifier}, in: \bibinfo{booktitle}{Proceedings of the 22nd ACM SIGKDD international conference on knowledge discovery and data mining}, pp. \bibinfo{pages}{1135--1144}.
%Type = Inproceedings
\bibitem[{Ronneberger et~al.(2015)Ronneberger, Fischer and Brox}]{ronneberger2015u}
\bibinfo{author}{Ronneberger, O.}, \bibinfo{author}{Fischer, P.}, \bibinfo{author}{Brox, T.}, \bibinfo{year}{2015}.
\newblock \bibinfo{title}{U-net: Convolutional networks for biomedical image segmentation}, in: \bibinfo{booktitle}{Medical Image Computing and Computer-Assisted Intervention--MICCAI 2015: 18th International Conference, Munich, Germany, October 5-9, 2015, Proceedings, Part III 18}, \bibinfo{organization}{Springer}. pp. \bibinfo{pages}{234--241}.
%Type = Article
\bibitem[{Rudin(2019)}]{rudin2019stop}
\bibinfo{author}{Rudin, C.}, \bibinfo{year}{2019}.
\newblock \bibinfo{title}{Stop explaining black box machine learning models for high stakes decisions and use interpretable models instead}.
\newblock \bibinfo{journal}{Nature Machine Intelligence} \bibinfo{volume}{1}, \bibinfo{pages}{206--215}.
%Type = Article
\bibitem[{R\u{u}{\v{z}}i{\v{c}}ka et~al.(2022)R\u{u}{\v{z}}i{\v{c}}ka, Vaughan, De~Martini, Fulton, Salvatelli, Bridges, Mateo-Garcia and Zantedeschi}]{ruuvzivcka2022ravaen}
\bibinfo{author}{R\u{u}{\v{z}}i{\v{c}}ka, V.}, \bibinfo{author}{Vaughan, A.}, \bibinfo{author}{De~Martini, D.}, \bibinfo{author}{Fulton, J.}, \bibinfo{author}{Salvatelli, V.}, \bibinfo{author}{Bridges, C.}, \bibinfo{author}{Mateo-Garcia, G.}, \bibinfo{author}{Zantedeschi, V.}, \bibinfo{year}{2022}.
\newblock \bibinfo{title}{Rav{\ae}n: unsupervised change detection of extreme events using ml on-board satellites}.
\newblock \bibinfo{journal}{Scientific reports} \bibinfo{volume}{12}, \bibinfo{pages}{16939}.
%Type = Article
\bibitem[{Schlaffer et~al.(2015)Schlaffer, Matgen, Hollaus and Wagner}]{schlaffer2015flood}
\bibinfo{author}{Schlaffer, S.}, \bibinfo{author}{Matgen, P.}, \bibinfo{author}{Hollaus, M.}, \bibinfo{author}{Wagner, W.}, \bibinfo{year}{2015}.
\newblock \bibinfo{title}{Flood detection from multi-temporal sar data using harmonic analysis and change detection}.
\newblock \bibinfo{journal}{International Journal of Applied Earth Observation and Geoinformation} \bibinfo{volume}{38}, \bibinfo{pages}{15--24}.
%Type = Inproceedings
\bibitem[{Sculley(2010)}]{sculley2010web}
\bibinfo{author}{Sculley, D.}, \bibinfo{year}{2010}.
\newblock \bibinfo{title}{Web-scale k-means clustering}, in: \bibinfo{booktitle}{Proceedings of the 19th international conference on World wide web}, pp. \bibinfo{pages}{1177--1178}.
%Type = Inproceedings
\bibitem[{Selvaraju et~al.(2017)Selvaraju, Cogswell, Das, Vedantam, Parikh and Batra}]{selvaraju2017grad}
\bibinfo{author}{Selvaraju, R.R.}, \bibinfo{author}{Cogswell, M.}, \bibinfo{author}{Das, A.}, \bibinfo{author}{Vedantam, R.}, \bibinfo{author}{Parikh, D.}, \bibinfo{author}{Batra, D.}, \bibinfo{year}{2017}.
\newblock \bibinfo{title}{Grad-cam: Visual explanations from deep networks via gradient-based localization}, in: \bibinfo{booktitle}{Proceedings of the IEEE international conference on computer vision}, pp. \bibinfo{pages}{618--626}.
%Type = Inproceedings
\bibitem[{Strudel et~al.(2021)Strudel, Garcia, Laptev and Schmid}]{strudel2021segmenter}
\bibinfo{author}{Strudel, R.}, \bibinfo{author}{Garcia, R.}, \bibinfo{author}{Laptev, I.}, \bibinfo{author}{Schmid, C.}, \bibinfo{year}{2021}.
\newblock \bibinfo{title}{Segmenter: Transformer for semantic segmentation}, in: \bibinfo{booktitle}{Proceedings of the IEEE/CVF international conference on computer vision}, pp. \bibinfo{pages}{7262--7272}.
%Type = Article
\bibitem[{Tellman et~al.(2021)Tellman, Sullivan, Kuhn, Kettner, Doyle, Brakenridge, Erickson and Slayback}]{tellman2021satellite}
\bibinfo{author}{Tellman, B.}, \bibinfo{author}{Sullivan, J.}, \bibinfo{author}{Kuhn, C.}, \bibinfo{author}{Kettner, A.}, \bibinfo{author}{Doyle, C.}, \bibinfo{author}{Brakenridge, G.}, \bibinfo{author}{Erickson, T.}, \bibinfo{author}{Slayback, D.}, \bibinfo{year}{2021}.
\newblock \bibinfo{title}{Satellite imaging reveals increased proportion of population exposed to floods}.
\newblock \bibinfo{journal}{Nature} \bibinfo{volume}{596}, \bibinfo{pages}{80--86}.
%Type = Article
\bibitem[{Wang et~al.(2022a)Wang, Han, Zhou and Liu}]{wang2022visual}
\bibinfo{author}{Wang, W.}, \bibinfo{author}{Han, C.}, \bibinfo{author}{Zhou, T.}, \bibinfo{author}{Liu, D.}, \bibinfo{year}{2022}a.
\newblock \bibinfo{title}{Visual recognition with deep nearest centroids}.
\newblock \bibinfo{journal}{arXiv preprint arXiv:2209.07383} .
%Type = Article
\bibitem[{Wang et~al.(2022b)Wang, Braham, Xiong, Liu, Albrecht and Zhu}]{wang2022ssl4eo}
\bibinfo{author}{Wang, Y.}, \bibinfo{author}{Braham, N.A.A.}, \bibinfo{author}{Xiong, Z.}, \bibinfo{author}{Liu, C.}, \bibinfo{author}{Albrecht, C.M.}, \bibinfo{author}{Zhu, X.X.}, \bibinfo{year}{2022}b.
\newblock \bibinfo{title}{Ssl4eo-s12: A large-scale multi-modal, multi-temporal dataset for self-supervised learning in earth observation}.
\newblock \bibinfo{journal}{arXiv preprint arXiv:2211.07044} .
%Type = Article
\bibitem[{Wu et~al.(2022)Wu, Song, Huang, Zhong, Zhan, Teng, Qiu, He and Cao}]{wu2022flood}
\bibinfo{author}{Wu, H.}, \bibinfo{author}{Song, H.}, \bibinfo{author}{Huang, J.}, \bibinfo{author}{Zhong, H.}, \bibinfo{author}{Zhan, R.}, \bibinfo{author}{Teng, X.}, \bibinfo{author}{Qiu, Z.}, \bibinfo{author}{He, M.}, \bibinfo{author}{Cao, J.}, \bibinfo{year}{2022}.
\newblock \bibinfo{title}{Flood detection in dual-polarization sar images based on multi-scale deeplab model}.
\newblock \bibinfo{journal}{Remote Sensing} \bibinfo{volume}{14}, \bibinfo{pages}{5181}.
%Type = Article
\bibitem[{Xie et~al.(2021)Xie, Wang, Yu, Anandkumar, Alvarez and Luo}]{xie2021segformer}
\bibinfo{author}{Xie, E.}, \bibinfo{author}{Wang, W.}, \bibinfo{author}{Yu, Z.}, \bibinfo{author}{Anandkumar, A.}, \bibinfo{author}{Alvarez, J.M.}, \bibinfo{author}{Luo, P.}, \bibinfo{year}{2021}.
\newblock \bibinfo{title}{Segformer: Simple and efficient design for semantic segmentation with transformers}.
\newblock \bibinfo{journal}{Advances in Neural Information Processing Systems} \bibinfo{volume}{34}, \bibinfo{pages}{12077--12090}.
%Type = Article
\bibitem[{Xu(2006)}]{xu2006modification}
\bibinfo{author}{Xu, H.}, \bibinfo{year}{2006}.
\newblock \bibinfo{title}{Modification of normalised difference water index (ndwi) to enhance open water features in remotely sensed imagery}.
\newblock \bibinfo{journal}{International journal of remote sensing} \bibinfo{volume}{27}, \bibinfo{pages}{3025--3033}.
%Type = Article
\bibitem[{Yan et~al.(2019)Yan, Wang, Song, Chen, Chen and Deng}]{yan2019time}
\bibinfo{author}{Yan, J.}, \bibinfo{author}{Wang, L.}, \bibinfo{author}{Song, W.}, \bibinfo{author}{Chen, Y.}, \bibinfo{author}{Chen, X.}, \bibinfo{author}{Deng, Z.}, \bibinfo{year}{2019}.
\newblock \bibinfo{title}{A time-series classification approach based on change detection for rapid land cover mapping}.
\newblock \bibinfo{journal}{ISPRS Journal of Photogrammetry and Remote Sensing} \bibinfo{volume}{158}, \bibinfo{pages}{249--262}.
%Type = Article
\bibitem[{Zhang et~al.(2021)Zhang, Zhang and Hu}]{zhang2021unsupervised}
\bibinfo{author}{Zhang, Q.}, \bibinfo{author}{Zhang, P.}, \bibinfo{author}{Hu, X.}, \bibinfo{year}{2021}.
\newblock \bibinfo{title}{Unsupervised grnn flood mapping approach combined with uncertainty analysis using bi-temporal sentinel-2 msi imageries}.
\newblock \bibinfo{journal}{International Journal of Digital Earth} \bibinfo{volume}{14}, \bibinfo{pages}{1561--1581}.
%Type = Article
\bibitem[{Zhang et~al.(2022)Zhang, Angelov, Soares, Longepe and Mathieu}]{zhang2022interpretable}
\bibinfo{author}{Zhang, Z.}, \bibinfo{author}{Angelov, P.}, \bibinfo{author}{Soares, E.}, \bibinfo{author}{Longepe, N.}, \bibinfo{author}{Mathieu, P.P.}, \bibinfo{year}{2022}.
\newblock \bibinfo{title}{An interpretable deep semantic segmentation method for earth observation}.
\newblock \bibinfo{journal}{arXiv preprint arXiv:2210.12820} .
%Type = Article
\bibitem[{Zhao et~al.(2023)Zhao, Sui and Liu}]{zhao2023siam}
\bibinfo{author}{Zhao, B.}, \bibinfo{author}{Sui, H.}, \bibinfo{author}{Liu, J.}, \bibinfo{year}{2023}.
\newblock \bibinfo{title}{Siam-dwenet: Flood inundation detection for sar imagery using a cross-task transfer siamese network}.
\newblock \bibinfo{journal}{International Journal of Applied Earth Observation and Geoinformation} \bibinfo{volume}{116}, \bibinfo{pages}{103132}.
%Type = Inproceedings
\bibitem[{Zheng et~al.(2021)Zheng, Lu, Zhao, Zhu, Luo, Wang, Fu, Feng, Xiang, Torr et~al.}]{zheng2021rethinking}
\bibinfo{author}{Zheng, S.}, \bibinfo{author}{Lu, J.}, \bibinfo{author}{Zhao, H.}, \bibinfo{author}{Zhu, X.}, \bibinfo{author}{Luo, Z.}, \bibinfo{author}{Wang, Y.}, \bibinfo{author}{Fu, Y.}, \bibinfo{author}{Feng, J.}, \bibinfo{author}{Xiang, T.}, \bibinfo{author}{Torr, P.H.}, et~al., \bibinfo{year}{2021}.
\newblock \bibinfo{title}{Rethinking semantic segmentation from a sequence-to-sequence perspective with transformers}, in: \bibinfo{booktitle}{Proceedings of the IEEE/CVF conference on computer vision and pattern recognition}, pp. \bibinfo{pages}{6881--6890}.
%Type = Article
\bibitem[{Zheng et~al.(2018)Zheng, Lei, Yao, Gong and Yin}]{zheng2018image}
\bibinfo{author}{Zheng, X.}, \bibinfo{author}{Lei, Q.}, \bibinfo{author}{Yao, R.}, \bibinfo{author}{Gong, Y.}, \bibinfo{author}{Yin, Q.}, \bibinfo{year}{2018}.
\newblock \bibinfo{title}{Image segmentation based on adaptive k-means algorithm}.
\newblock \bibinfo{journal}{EURASIP Journal on Image and Video Processing} \bibinfo{volume}{2018}, \bibinfo{pages}{1--10}.
%Type = Article
\bibitem[{Zhou et~al.(2017)Zhou, Dong, Xiao, Xiao, Yang, Zhao, Zou and Qin}]{zhou2017open}
\bibinfo{author}{Zhou, Y.}, \bibinfo{author}{Dong, J.}, \bibinfo{author}{Xiao, X.}, \bibinfo{author}{Xiao, T.}, \bibinfo{author}{Yang, Z.}, \bibinfo{author}{Zhao, G.}, \bibinfo{author}{Zou, Z.}, \bibinfo{author}{Qin, Y.}, \bibinfo{year}{2017}.
\newblock \bibinfo{title}{Open surface water mapping algorithms: A comparison of water-related spectral indices and sensors}.
\newblock \bibinfo{journal}{Water} \bibinfo{volume}{9}, \bibinfo{pages}{256}.
%Type = Article
\bibitem[{Zhou et~al.(2022)Zhou, Yang, Ma, Hu and Zhang}]{zhou2022water}
\bibinfo{author}{Zhou, Y.}, \bibinfo{author}{Yang, K.}, \bibinfo{author}{Ma, F.}, \bibinfo{author}{Hu, W.}, \bibinfo{author}{Zhang, F.}, \bibinfo{year}{2022}.
\newblock \bibinfo{title}{Water--land segmentation via structure-aware cnn--transformer network on large-scale sar data}.
\newblock \bibinfo{journal}{IEEE Sensors Journal} \bibinfo{volume}{23}, \bibinfo{pages}{1408--1422}.
%Type = Article
\bibitem[{Zhu(2017)}]{zhu2017change}
\bibinfo{author}{Zhu, Z.}, \bibinfo{year}{2017}.
\newblock \bibinfo{title}{Change detection using landsat time series: A review of frequencies, preprocessing, algorithms, and applications}.
\newblock \bibinfo{journal}{ISPRS Journal of Photogrammetry and Remote Sensing} \bibinfo{volume}{130}, \bibinfo{pages}{370--384}.

\end{thebibliography}

\appendix
\setcounter{equation}{0}
\renewcommand{\theequation}{\arabic{equation}}

\section*{APPENDIX}\label{APPENDIX}

The expression of similarity, $s$, which follows the density expression from \citep{angelov2012autonomous} is shown below:

\begin{equation}
    s(x_{h,v}^k )=\frac{1}{(1+||x_{h,v}^k-\mu_{h,v}^{k}\||^2+\Sigma_{h,v}^k-||\mu_{h,v}^{k}||^2 )}
\end{equation}
Here, the mean value $\mu_{h,v}$ and the scalar product $\Sigma_{h,v}^k$ can be updated recursively as follows \citep{angelov2012autonomous}:
\begin{equation}
    \mu_{h,v}^{k} = \frac{1}{L}\sum_{l=1}^{L}x_{h,v}^{L}
\end{equation}
\begin{equation}
    \Sigma_{h,v}^{K} = \frac{k-1}{k}\Sigma_{h,v}^{k-1}+\frac{1}{k}||x_{h,v}^k||^2
\end{equation}
Here, L represents the number of images that do not identify as novel images.

Furthermore, the overall image similarity, $S^k$, is calculated by taking the mean value of the pixel-wise similarity of the $k$-th image using Equation (4) \citep{angelov2012autonomous}:
\begin{equation}
    S^k = \frac{1}{nm}\sum_{i=1}^{n}\sum_{j=1}^{m}s(x_{h,v}^{k})
\end{equation}

Subsequently, the mean similarity $\Bar{S}$ and standard deviation $\sigma$ are updated recursively as follows \citep{angelov2012autonomous}:

\begin{equation}
    \Bar{S}^k=\frac{k-1}{k}\Bar{S}^k-1+\frac{1}{k}S^k,  \Bar{S}^1=S^1
\end{equation}

\begin{equation}
    (\sigma^k)^2 = \frac{k-1}{k}(\sigma^{k-1})^2+\frac{1}{k}(S^k-\Bar{S}^k)^2,  (\sigma^1)^2=0
\end{equation}

% \end{thebibliography}
\end{document}